\documentclass[]{article}

\usepackage[final]{neurips_2022}
\usepackage[utf8]{inputenc} 
\usepackage[T1]{fontenc}    
\usepackage{hyperref}       
\usepackage{booktabs}       
\usepackage{amsfonts}       
\usepackage{amsmath}
\usepackage{nicefrac}       
\usepackage{microtype}      
\usepackage{color}         
\usepackage{xcolor}         
\usepackage{graphicx}
\usepackage{graphics}
\usepackage{caption}
\usepackage{subcaption}
\usepackage[linesnumbered, ruled,vlined]{algorithm2e}
\usepackage{algpseudocode}
\usepackage{multirow}
\usepackage{xr}
\usepackage{soul}
\usepackage{makecell}
\usepackage{enumitem}
\usepackage{bbding}

\newcommand{\name}{CodeGeeX\xspace}
\newcommand{\bench}{HumanEval-X\xspace}
\newcommand{\vpara}[1]{\vspace{0.07in}\noindent\textbf{#1}\xspace}

\newcommand{\hide}[1]{} 

\title{\name: A Pre-Trained Model for Code Generation with Multilingual Benchmarking on HumanEval-X}

\author{\\
\bf Qinkai Zheng$^{\mathsection\circ}$$^*$, Xiao Xia$^{\mathsection}$\thanks{QZ and XX contributed equally. Emails: \texttt{\{qinkai|xiax19\}@tsinghua.edu.cn}} , Xu Zou$^{\mathsection}$, Yuxiao Dong$^{\mathsection}$\thanks{Team Leads: YD, ZY, and JT. Emails: \texttt{\{yuxiaod|zhiliny|jietang\}@tsinghua.edu.cn}} , Shan Wang$^{\circ}$, Yufei Xue$^{\circ}$, Zihan Wang$^{\mathsection}$, \\ 
\bf Lei Shen$^{\circ}$, Andi Wang$^{\circ}$, Yang Li$^{\circ}$, Teng Su$^{\diamond}$, Zhilin Yang$^{\mathsection\dagger}$, Jie Tang$^{\mathsection\dagger}$\thanks{Corresponding author: JT. Email: \texttt{jietang@tsinghua.edu.cn}} \\ \\
Tsinghua University$^{\mathsection}$, Zhipu.AI$^{\circ}$, Huawei$^{\diamond}$ 
}

\begin{document}

\maketitle

\begin{abstract}
    Large pre-trained code generation models, such as OpenAI Codex, can generate syntax- and function-correct code, making the coding of programmers more productive and our pursuit of artificial general intelligence closer. 
    In this paper, we introduce \name, a multilingual model with 13 billion parameters for code generation. 
    \name is pre-trained on 850 billion tokens of 23 programming languages as of June 2022. 
    Our extensive experiments suggest that \name outperforms multilingual code models of similar scale for both the tasks of code generation and translation on \bench. 
    Building upon HumanEval (Python only), we develop the \bench benchmark for evaluating multilingual models by hand-writing the solutions in C++, Java, JavaScript, and Go. 
    In addition, we build \name-based extensions on Visual Studio Code, JetBrains, and Cloud Studio, generating 4.7 billion tokens for tens of thousands of active users per week. 
    Our user study demonstrates that \name can help to increase coding efficiency for 83.4\% of its users. 
    Finally, \name is publicly accessible and in Sep. 2022, we open-sourced its code, model weights (the version of 850B tokens), API, extensions, and \bench at \url{https://github.com/THUDM/CodeGeeX}.     
\end{abstract}

\section{Introduction}
Given the description of a human intent, such as \texttt{``write a factorial function''}, can the machine automatically generate an executable program that addresses this need? 
This is the problem of \textit{automatic program writing} that has been explored  since the early days of computer science in the 1960s~\citep{waldinger1969prow, summers1977methodology}. 
From LISP-based pioneering deductive synthesis approaches~\citep{waldinger1969prow, summers1977methodology} to modern program synthesis systems~\citep{solar2008program, polozov2015flashmeta}, to end-to-end code generation via deep neural networks~\citep{mou2015end,svyatkovskiy2020intellicode,sun2020treegen}, tremendous efforts have been made to enable machines to automatically write correct programs as part of the quest to artificial general intelligence. 

By treating programs as language sequences, neural sequential architectures, such as recurrent neural networks and transformer~\citep{vaswani2017transformer}, can be naturally applied to code generation. 
In fact, transformer-based techniques~\citep{svyatkovskiy2020intellicode,sun2020treegen} have shown the potential of \textit{automatic program writing} by starting to generate code that is both syntactically correct and consistent in 2020.
This progress is significantly furthered when large language models (transformers with billions of parameters) meet the massive open-sourced code data. 

Notably, the OpenAI Codex~\citep{chen2021codex} model (Python only) with 12 billion (12B) parameters pioneered and demonstrated the potential of large code generation models pre-trained on billions lines of public code.
By using the generative pre-training (GPT) strategy, Codex can solve introductory-level programming problems in Python with a high probability. 
Research studies~\citep{ziegler2022productivity} also show that 88\% of users of GitHub Copilot---a paid service powered by Codex---feel more productive when coding with it. 
Since then, large pre-trained code models have been extensively developed, including DeepMind AlphaCode~\citep{li2022alphacode}, 
Salesforce CodeGen~\citep{nijkamp2022codegen}, 
Meta InCoder~\citep{fried2022incoder}, 
and Google PaLM-Coder-540B~\citep{chowdhery2022Palm}. 

In this work, we present \name, a multilingual code generation model with 13 billion parameters, pre-trained on a large code corpus of 23 programming languages. 
It was trained on more than 850 billion tokens on a cluster of 1,536 Ascend 910 AI Processors between April and June 2022, and was publicly released in Sep. 2022 (Cf. the GitHub repo).
\name has the following properties. 
First, different from Codex in \cite{chen2021codex}, both \name---the model itself---and how such scale of code models can be pre-trained are open-sourced, facilitating the understanding and advances in pre-trained code generation models. 
\name also supports cross-platform inference on both Ascend and NVIDIA GPUs.
Second, in addition to code generation and code completion as Codex and others, \name supports the tasks of code explanation and code translation between language pairs (Cf. Figure 1 (a)). 
Third, it offers consistent performance advantages over well-known \textit{multilingual} code generation models of the similar scale, including CodeGen-16B, GPT-NeoX-20B, InCode-6.7B, and GPT-J-6B (Cf. Figure 1 (b) and (c)).

\begin{figure*}
    \centering
    \includegraphics[width=\textwidth]{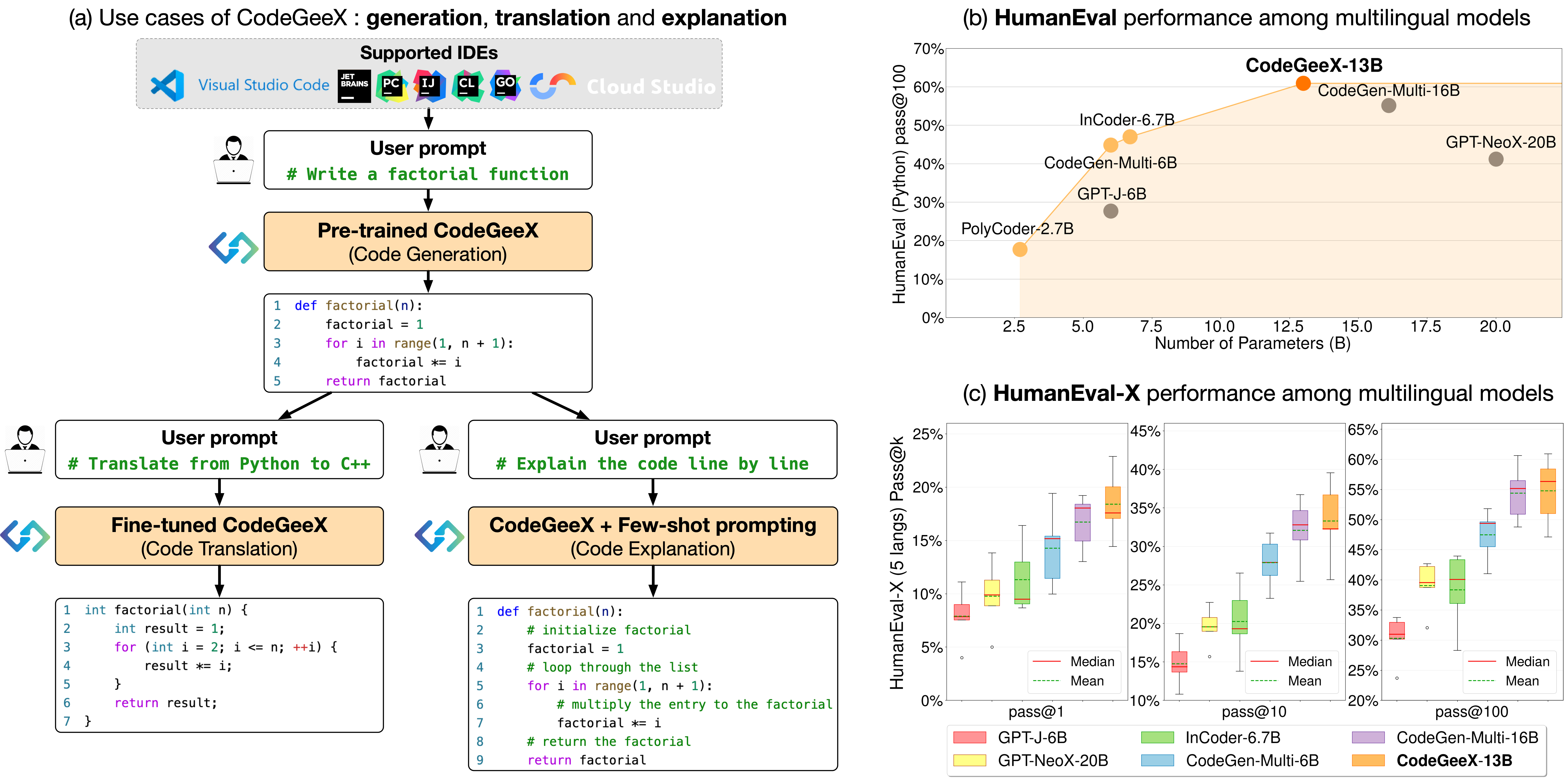}
    \caption{Summary of \name. 
    \textmd{(a): In supported IDEs, users can interact with \name by providing prompts. Different models are used to support three tasks: code generation, code translation and code explanation. (b) and (c): In HumanEval and our newly-proposed HumanEval-X, \name shows promising multilingual abilities and consistently outperforms other multilingual code generation models.}
    }
    \label{fig:codegeex_summary}
\end{figure*}

\begin{table*}
\renewcommand{\arraystretch}{1.3}
\centering
\caption{Large pre-trained language models related to programming languages in the literature.}
\label{tab:code_models}
\resizebox{\textwidth}{!}{
\begin{tabular}{|c|c|c|c|c|c|c|c|c|c|c|} 
\toprule
 & \multicolumn{3}{c|}{\textbf{Model Properties}} & \multicolumn{3}{c|}{\textbf{Dataset}} & \multicolumn{3}{c|}{\textbf{Evaluation}} \\ 
\hline
 & Open & \begin{tabular}[c]{@{}c@{}}Multi-\\lingual\end{tabular} & \# Params & Source & Languages & Size & \begin{tabular}[c]{@{}c@{}}Multilingual\\Evaluation\end{tabular} & Translation & Benchmark \\ 
\hline
Codex~\citep{chen2021codex} & \XSolidBrush & \XSolidBrush & 12B & Collected & Python & Code: 159GB & \XSolidBrush & \XSolidBrush & HumanEval, APPS  \\ 
\hline
AlphaCode~\citep{li2022alphacode} & \XSolidBrush & \Checkmark & 41B & Collected & 12 langs & Code: 715.1GB & \Checkmark & \XSolidBrush & \begin{tabular}[c]{@{}c@{}}HumanEval, APPS\\CodeContest\end{tabular}  \\
\hline
PaLM-Coder~\citep{chowdhery2022Palm} & \XSolidBrush & \Checkmark & 8B, 62B, 540B & Collected & Multiple & \begin{tabular}[c]{@{}c@{}}Text: 741B tokens\\Code: 39GB\\(780B tokens trained)\end{tabular} & \Checkmark & \Checkmark & \begin{tabular}[c]{@{}c@{}}HumanEval, MBPP\\TransCoder, DeepFix\end{tabular}  \\
\hline
PolyCoder~\citep{xu2022polycoder} & \Checkmark & \Checkmark & 2.7B & Collected & 12 langs & Code: 253.6GB & \XSolidBrush & \XSolidBrush & HumanEval  \\ 
\hline
GPT-Neo~\citep{black2021gptneo} & \Checkmark & \Checkmark & 1.3B, 2.7B & The Pile & Multiple & \begin{tabular}[c]{@{}c@{}}Text: 730GB\\Code: 96GB\\(400B tokens trained)\end{tabular} & \XSolidBrush & \XSolidBrush & HumanEval \\ 
\hline
GPT-NeoX~\citep{black2022gptneox} & \Checkmark & \Checkmark & 20B & The Pile & Multiple & \begin{tabular}[c]{@{}c@{}}Text: 730GB\\Code: 95GB\\(473B tokens trained)\end{tabular} & \XSolidBrush & \XSolidBrush & HumanEval \\ 
\hline
GPT-J~\citep{wang2021gptj} & \Checkmark & \Checkmark & 6B & The Pile & Multiple & \begin{tabular}[c]{@{}c@{}}Text: 730GB\\Code: 96GB\\(473B tokens trained)\end{tabular} & \XSolidBrush & \XSolidBrush & HumanEval \\
\hline
Incoder~\citep{fried2022incoder} & \Checkmark & \Checkmark & 1.3B, 6.7B & Collected & 28 langs & \begin{tabular}[c]{@{}c@{}}Code: 159GB\\StackOverflow: 57GB\\(60B tokens trained)\end{tabular} & \XSolidBrush & \XSolidBrush & \begin{tabular}[c]{@{}c@{}}HumanEval, MBPP\\CodeXGLUE\end{tabular} \\ 
\hline
CodeGen-Multi~\citep{nijkamp2022codegen} & \Checkmark & \Checkmark & 6.1B, 16.1B & BigQuery & 6 langs & \begin{tabular}[c]{@{}c@{}}Code: 150B tokens\\Text: 355B tokens\\(1000B tokens trained)\end{tabular} & \XSolidBrush & \XSolidBrush & HumanEval, MTPB \\ 
\hline
CodeGen-Mono~\citep{nijkamp2022codegen} & \Checkmark & \XSolidBrush & 6.1B, 16.1B & BigPython & Python & \begin{tabular}[c]{@{}c@{}}Code: 150B tokens\\Text: 355B tokens\\(1300B tokens trained)\end{tabular} & \XSolidBrush & \XSolidBrush & HumanEval, MTPB \\ 
\hline
\textbf{CodeGeeX} & \color{orange}\Checkmark & \color{orange}\Checkmark & 13B & \begin{tabular}[c]{@{}c@{}}The Pile\\CodeParrot\\Collected\end{tabular} & 23 langs & \begin{tabular}[c]{@{}c@{}}Code: 158B tokens\\(850B tokens trained)\end{tabular} & \color{orange}\Checkmark & \color{orange}\Checkmark & \begin{tabular}[c]{@{}c@{}}\textcolor{orange}{\bench}, HumanEval\\MBPP, CodeXGLUE, XLCoST\end{tabular} \\
\bottomrule
\end{tabular}}
\end{table*}

We also build the free \name extensions in several IDEs, currently including Visual Studio Code, JetBrains, and Tencent Cloud Studio (a Web IDE). 
It supports several different modes---code completion, function-level generation, code translation, code explanation, and customizable prompting---to help users' programming tasks in real time. 
Since its release, there are tens of thousands of daily active users, each of which on average makes 250+ API calls per weekday. 
As of this writing, the \name model generates 4.7 billion tokens per week. 
Our user survey suggests that 83.4\% of users feel the \name extensions improve their programming efficiency.

Finally, we develop the \bench benchmark for evaluating multilingual code models as 1) HumanEval~\citep{chen2021codex}---developed by OpenAI for evaluating Codex---and other benchmarks~\citep{austin2021program,hendrycks2021apps, nijkamp2022codegen} only consist of programming problems in a single language  
and 
2) existing multilingual datasets~\citep{ren2020codebleu,lu2021codexglue,zhu2022xlcost} use string similarity metrics like BLEU~\citep{papineni2002bleu} for evaluation rather than really verify the functional correctness of generated code. 
Specifically, for each problem---defined only for Python---in HumanEval, we manually rewrite its prompt, canonical solution, and test cases in C++, Java, JavaScript, and Go.
In total, \bench covers 820 hand-written problem-solution pairs (164 problems, each having solutions in 5 languages). 
Importantly, \bench support the evaluation of both code generation and code translation between different languages. 

The  contributions of this work can be summarized as follows:
\begin{itemize}[leftmargin=*]
    \item We develop and release \name, a 13B pre-trained 23-language code generation model that demonstrates consistent outperformance on code generation and translation over its multilingual baselines of the same scale. 
    
    \item We build the \name extensions on VS Code\footnote{\url{https://marketplace.visualstudio.com/items?itemName=aminer.codegeex}}, JebBrains\footnote{\url{https://plugins.jetbrains.com/plugin/20587-codegeex}}, and Tencent Cloud Studio. 
    Compared to Copilot, it supports more diverse functions, including code completion, generation, translation, and explanation. 
    According to the user survey,  \name can improve the coding efficiency for 83.4\% of its users.  
    
    \item We hand-craft the \bench benchmark to evaluate multilingual code models for the tasks of code generation and translation in terms of functional correctness, facilitating the understanding and development of pre-trained (multilingual) code models. 
\end{itemize}

\label{sec:intro}

\section{The CodeGeeX Model}
\name is a multilingual code generation model with 13 billion (13B) parameters, pre-trained on a large code corpus of 23 programming languages. 
As of June 22, 2022, \name has been trained on more than 850 billion tokens on a cluster of 1,536 Ascend 910 AI Processors for over two months.

We introduce the \name model and its design choices. 
The consensus reality is that it is computationally unaffordable to test different architectural designs for large pre-trained models~\citep{brown2020gpt3, chowdhery2022Palm, zhang2022opt, zeng2022glm}, though they define the inductive bias of models.

\subsection{\name's Architecture}

\begin{figure}[htbp]
    \centering
    \includegraphics[width=0.7\columnwidth]{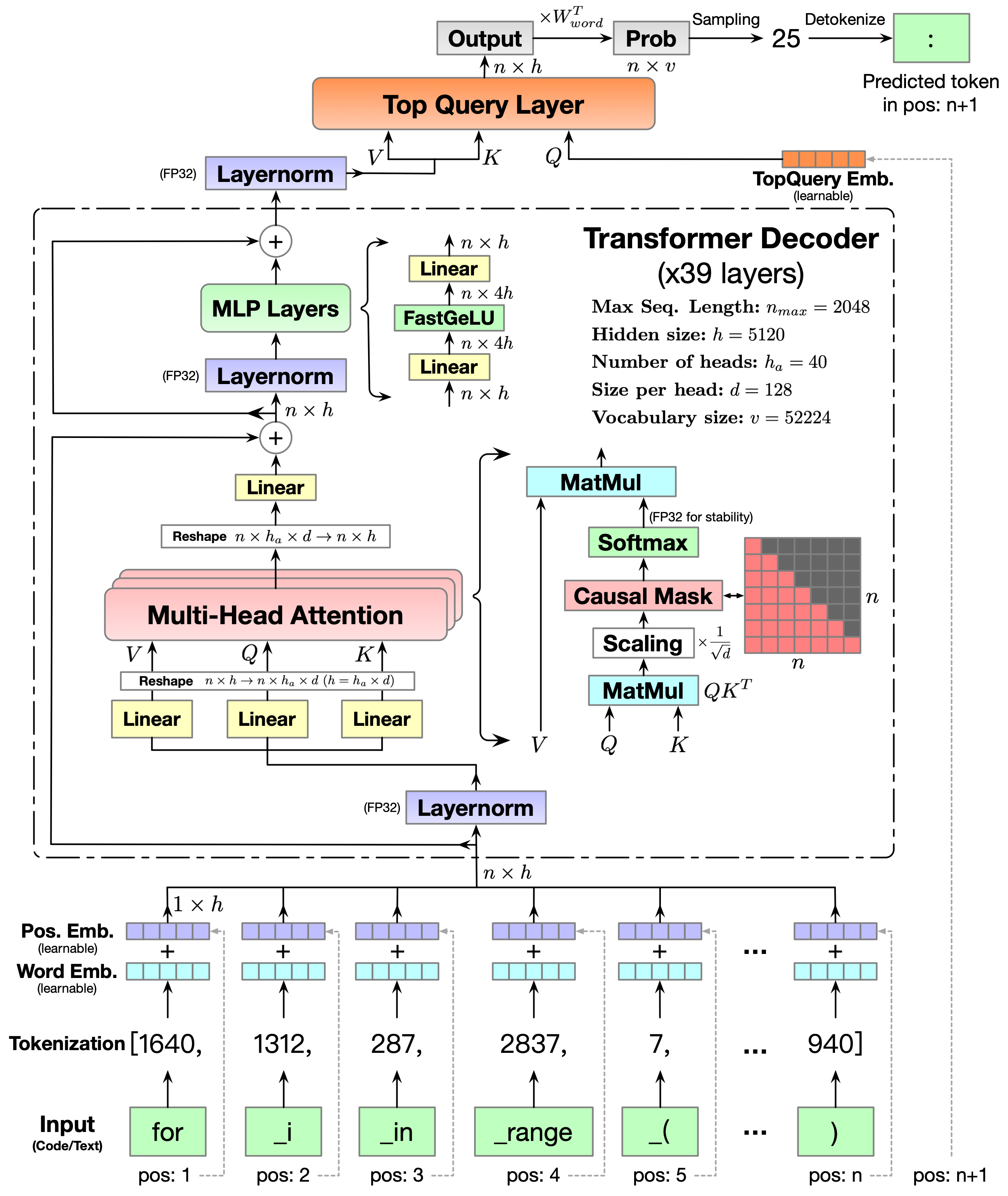}
    \caption{\name's model architecture. 
    \textmd{\name is a code generation model with 13B parameters,  consisting of 39-layer left-to-right transformer decoders and a top query layer. It takes text/code tokens as input and outputs the probability of the next token autoregressively.}
    }
    \label{fig:arch}
\end{figure}

\vpara{The Transformer Backbone.}
Similar to recent pre-trained models, such as GPT-3~\citep{brown2020gpt3}, PaLM~\citep{chowdhery2022Palm}, and Codex~\citep{chen2021codex}, \name follows the generative pre-training (GPT) architecture~\citep{radford2018gpt} with the decoder-only style for autoregressive (programming) language modeling. 
The core architecture of \name is a 39-layer transformer decoder.
In each transformer layer (in~\figurename~\ref{fig:arch}), we apply a multi-head self-attention mechanism~\citep{vaswani2017transformer} followed by MLP layers, together with layer normalization~\citep{ba2016layernorm} and residual connection~\citep{he2016resnet}. 
We use an approximation of GELU (Gaussian Linear Units) operation~\citep{hendrycks2016gelu}, namely FastGELU, which is more efficient under the Ascend 910 AI Processor:
\begin{equation}
\small
    \text{FastGELU}(X_i)=\frac{X_i}{1+\exp(-1.702*|X_i|)*\exp(0.851*(X_i-|X_i|))}
\end{equation}

\vpara{Generative Pre-Training Objective.}
By adopting the GPT paradigm~\citep{radford2019gpt2, chen2021codex}, we train the model on a large amount of unlabeled code data.
The principle is to iteratively take code tokens as input, predict the next token, and compare it with the ground truth.
Specifically, for any input sequence $\{x_1, x_2, ..., x_n\}$ of length $n$, the output of \name is a probability distribution of the next token $\mathbb{P}(x_{n+1}|x_1, x_2, ..., x_{n}, \Theta)=p_{n+1}\in[0,1]^{1\times v}$, where $\Theta$ represents all parameters of the model and $v$ is the vocabulary size.
By comparing it with the real distribution, \emph{i.e.}, a one-hot vector $y_{n+1}\in\{0,1\}^{1\times v}$ of the ground-truth token, we can optimize the cumulative cross-entropy loss:
\begin{equation}
\small
	\mathcal{L}=-\sum_{n=1}^{N-1} y_{n+1}\log \mathbb{P}(x_{n+1}|x_1, x_2, ..., x_{n}, \Theta)
\label{eq:cross_entropy}
\end{equation}

\vpara{The Top Query Layer and Decoding.} 
The original GPT model uses a pooler function to obtain the final output. 
We use an extra query layer~\citep{zeng2021pangu-alpha} on top of all other transformer layers to obtain the final embedding through attention.
As shown in~\figurename~\ref{fig:arch}, the input of the top query layer replaces the query input $X_{in}$ by the query embedding of position $n+1$. 
The final output is multiplied by the transpose of word embedding matrix to get the output probability. 
For decoding strategies, \name supports greedy, temperature sampling, top-k sampling, top-p sampling, and beam search. 
Finally, detokenization will turn the selected token ID into an actual word.

\begin{figure}[htbp]
    \centering
    \includegraphics[width=.93\columnwidth]{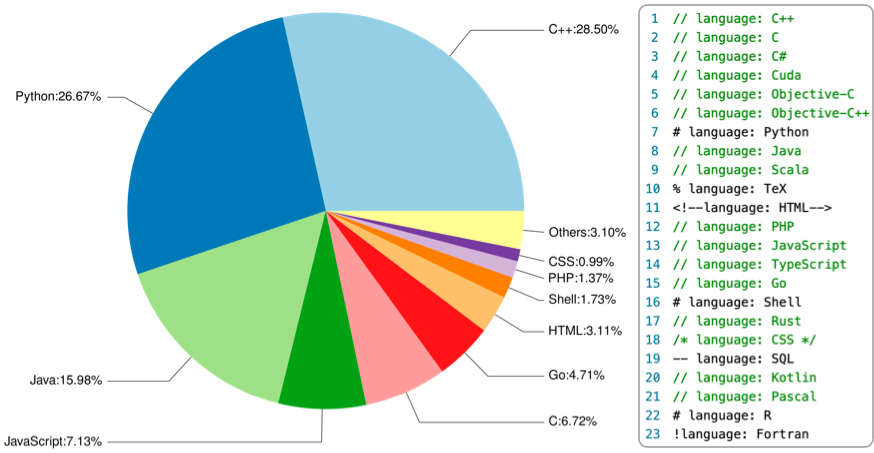}
    \caption{Language distribution and tags of \name's data.}
    \label{fig:data_pie}
\end{figure}

\subsection{Pre-Training Setup}

\vpara{Code Corpus.}
The training corpus contains two parts. 
The first part is from open source code datasets, the Pile~\citep{gao2020pile} and CodeParrot\footnote{\url{https://huggingface.co/datasets/transformersbook/codeparrot}}.
The Pile contains a subset of public repositories with more than 100 stars on GitHub,
from which we select files of 23 popular programming languages including C++, Python, Java, JavaScript, C, Go, and so on. 
We identify the programming language of each file based on its suffix and the major language of the repository it belongs to. 
CodeParrot is another public Python dataset from BigQuery. 
The second part is supplementary data of Python, Java, and C++ directly scraped from GitHub public repositories that do not appear in the first part. 
We choose repositories that have at least one star and a total size within 10MB, then we filter out files that: 1) have more than 100 characters per line on average, 2) are automatically generated, 3) have a ratio of alphabet less than 40\%, 4) are bigger than 100KB or smaller than 1KB. 
We format Python code according to the PEP8 standards.

\figurename~\ref{fig:data_pie} shows the composition of the 158B-token training data, containing 23 programming languages.
We divide the training data into segments of equal length. 
To help the model distinguish between multiple languages, we add a language-specific tag before each segment in the form of \texttt{[Comment sign]language:\ [LANG]}, \emph{e.g.}, \texttt{\# language:\ Python}.

\vpara{Tokenization.}
The first step is to convert code snippets into numerical vectors.
Considering that 1) there is a large number of natural language comments in code data, 2) the naming of variables, functions, and classes are often meaningful words, we treat code data the same as text data and apply the GPT-2 tokenizer~\citep{radford2019gpt2}. 
It is a BPE (Byte Pair Encoding)~\citep{sennrich2015bpe} tokenizer that deals with the open-vocabulary problem using a fixed-size vocabulary with variable-length characters.
The initial vocabulary size is 50,000, we encode multiple whitespaces as extra tokens following~\cite{chen2021codex} to increase the encoding efficiency. 
Specifically, \texttt{L} whitespaces are represented by \texttt{<|extratoken\_X|>}, where \texttt{X}=8+\texttt{L}.
Since the vocabulary contains tokens from various natural languages, it allows \name to process tokens in languages other than English, like Chinese, French, Russia, Japanese and more. 
The final vocabulary size is $v=52,224$. 
After tokenization, any code snippet or text description can be transformed into a vector of integers.
More details can be found in Appendix~\ref{app:tokenization}.

\vpara{The Input Word and Positional Embeddings.}
Given the tokens, the next step is to associate each token with a word embedding.
By looking up the token ID in a word embedding matrix $W_{word}\in\mathbb{R}^{v\times h}$, where 
$h=5120$ is the hidden size, a learnable embedding $x_{word}\in\mathbb{R}^h$ is obtained for each token. 
To capture positional information, we also adopt learnable positional embedding that maps the current position ID to a learnable embedding $x_{pos}\in\mathbb{R}^h$, from $W_{pos}\in\mathbb{R}^{n_{max}\times h}$, where $n_{max}=2048$ is the maximum sequence length. 
Then, two embeddings are added to obtain the input embeddings $x_{in}=x_{word}+x_{pos}$ for the model.
Finally, the entire sequence can be turned into input embeddings $X_{in}\in\mathbb{R}^{n\times h}$, where $n$ is the input sequence length.

\subsection{\name Training}

\vpara{Parallel Training on Ascend 910.}
\name was trained on a cluster of the Ascend 910 AI processors (32GB) with Mindspore (v1.7.0). 
We faced and addressed numerous unknown technical and engineering challenges during pre-training, as Ascend and Mindspore are relatively new compared to NVIDIA GPUs and PyTorch/TensorFlow. 
The entire pre-training process takes two months on 192 nodes with 1,536 AI processors, during which the model consumes 850B tokens, equivalent to 5+ epochs (213,000 steps). 
Detailed configurations can be found in~\tablename~\ref{tab:model-config}.

\begin{table}[htbp]
\centering
\caption{Training configuration of \name.}
\label{tab:model-config}
\resizebox{0.75\columnwidth}{!}{
\begin{tabular}{ccc} 
\toprule
\textbf{Category}                      & \textbf{Parameter}                             & \textbf{Value}                  \\
\midrule
\multirow{6}{*}{\textbf{Environment}}  & Framework                                      & Mindspore v1.7.0                \\
                                       & Hardwares                                      & 1,536x Ascend 910 AI processors  \\
                                       & Mem per GPU                                    & 32GB                            \\
                                       & GPUs per node                             & 8                               \\
                                       & CPUs per node                                   & 192                             \\
                                       & RAM per node                                   & 2048GB                          \\
\midrule
\multirow{12}{*}{\textbf{Model}}       & Model parameters                               & 13B                             \\
                                       & Vocabulary size                                & 52224                           \\
                                       & Position embedding                             & Learnable                       \\
                                       & Maximum sequence length                        & 2048                            \\
                                       & Hidden size $h$                                & 5120                            \\
                                       & Feed-forward size $4h$                         & 20480                           \\
                                       & Feed-forward activation                        & FastGELU                        \\
                                       & Layernorm epsilon                              & 1e-5                            \\
                                       & Layernorm precision                            & FP32                            \\
                                       & Number of attention heads $h_n$                & 40                              \\
                                       & Attention softmax precision                    & FP32                            \\
                                       & Dropout rate                                   & 0.1                             \\
\midrule
\multirow{3}{*}{\textbf{Parallelism}}  & Model parallel size                            & 8                               \\
                                       & Data parallel size                             & 192                             \\
                                       & Global batch size                              & 3072                            \\
\midrule
\multirow{9}{*}{\textbf{Optimization}} & Optimizer                                      & Adam                            \\
                                       & Optimizer parameters                           & $\beta_1=0.9,\beta_2=0.999$     \\
                                       & Initial/final learning rate                    & 1e-4/1e-6                           \\
                                       & Warm-up step                                   & 2000                            \\
                                       & Decay step                                     & 200000                          \\
                                       & Learning rate scheduler                       & cosine decay                     \\
                                       & Loss function $\mathcal{L}$                    & Cross entropy                   \\
                                       & Loss scaling                                   & Dynamic                         \\
                                       & Loss scaling window                            & 1000                            \\
                                       & Trained steps                                  & 213000
                                          \\
\bottomrule
\end{tabular}}
\end{table}

To increase training efficiency, we adopt an 8-way model parallel training together with 192-way data parallel training, with ZeRO-2~\citep{rajbhandari2020zero} optimizer enabled to further reduce the memory consumption of optimizer states. 
Finally, the micro-batch size is 16 per node and the global batch size reaches 3,072. 

Specifically, we use Adam optimizer~\citep{kingma2014adam} to optimize the loss in~\equationautorefname~\ref{eq:cross_entropy}. 
The model weights are under FP16 format, except that we use FP32 for layer-norm and softmax for higher precision and stability.
The model takes about 27GB of GPU memory.  
We start from an initial learning rate 1e-4, and apply a cosine learning rate decay by:
\begin{equation}
\small
 	lr_{current} = lr_{min} + 0.5 * (lr_{max} - lr_{min}) * (1 + \cos(\frac{n_{current}}{n_{decay}}\pi))
 	\label{eq:lr_schedular}
\end{equation}
During the two-month training, the training loss of \name continues to decrease as the training goes on. 
We evaluate the checkpoints on \bench code generation task and observe that the performance is continuously increasing. 
See details in Figures ~\ref{fig:training-loss} and \ref{fig:hx-iter} in Appendix~\ref{app:hx-additional}.


\vpara{Training Efficiency Optimization.}
Over the course of the training, we actively attempted to optimize the Mindspore framework to release the power of Ascend 910. 
Notably, we adopt the following techniques that significantly improve training efficiency:
\begin{itemize}[leftmargin=*]
    \item Kernel fusion: We fuse several element-wise operators to improve calculation efficiency on Ascend 910, including \texttt{Bias+LayerNorm}, \texttt{BatchMatmul+Add}, \texttt{FastGeLU+Matmul}, \texttt{Softmax}, etc.
    We also optimize \texttt{LayerNorm} operator to support multi-core calculation.
    \item Auto Tune optimization: 
    When loading models, Mindspore first compiles them to static computational graphs. 
    It uses the Auto Tune tool to optimize the choice of operators (\emph{e.g.}, matrix multiplication in different dimensions). And it applies graph optimization techniques to deal with operator fusion and constant folding.
\end{itemize}
\tablename~\ref{tab:compare_2_1} shows the comparison of training efficiency before and after our optimization. 
The overall efficiency is measured by trained tokens per day. 
We observe that the efficiency per processor was improved 3$\times$ compared to the non-optimized implementation and the overall token throughput of 1,536 GPUs was improved by 224\%.

\begin{table}[htb]
    \centering
    \caption{Training efficiency (before and after optimization).}
    \label{tab:compare_2_1}
    \resizebox{.75\columnwidth}{!}{
    \begin{tabular}{c|cc} 
    \toprule
     & \textbf{Before} & \textbf{After} \\
     \midrule
    \textbf{Device} & Ascend 910 & Ascend 910 \\
    \textbf{\#GPUs} & 1536 & 1536 \\
    \textbf{Parallelism} & Data parallel + Model parallel & Data parallel + Model parallel \\
    \textbf{Sequence length} & 2048 & 2048 \\
    \textbf{Global batch size} & 2048 & 3072 \\
    \textbf{Step time(s)} & 15s & 10s \\
    \textbf{Overall efficiency} & 24.2B tokens/day & 54.3B tokens/day \\
    \bottomrule
    \end{tabular}}
\end{table}

\subsection{Fast Inference}

To serve the pre-trained \name, we implement a pure PyTorch version of \name that supports inference on NVIDIA GPUs. 
To achieve fast and memory-efficient inference, we apply both quantization and acceleration techniques to the pre-trained \name.

\vpara{Quantization.} 
We apply post-training quantization techniques to decrease memory consumption of \name during inference. We transform weights $W$ in all linear transformations from FP16 to INT8 using the common absolute maximum quantization:
\begin{align}
\label{eq:quantization}
W_{q}=\text{Round}(\frac{W}{\lambda}), \lambda = \frac{\text{Max}(|W|)}{2^{b-1} - 1}
\end{align}
where $b$ is the bitwidth and $b=8$. $\lambda$ is the scaling factor. This quantization transform FP16 values in $[-\text{Max}(|W|), \text{Max}(|W|)]$ to integers between $[-127, 127]$. 

\begin{figure}[htbp]
    \centering
    \includegraphics[width=0.5\columnwidth]{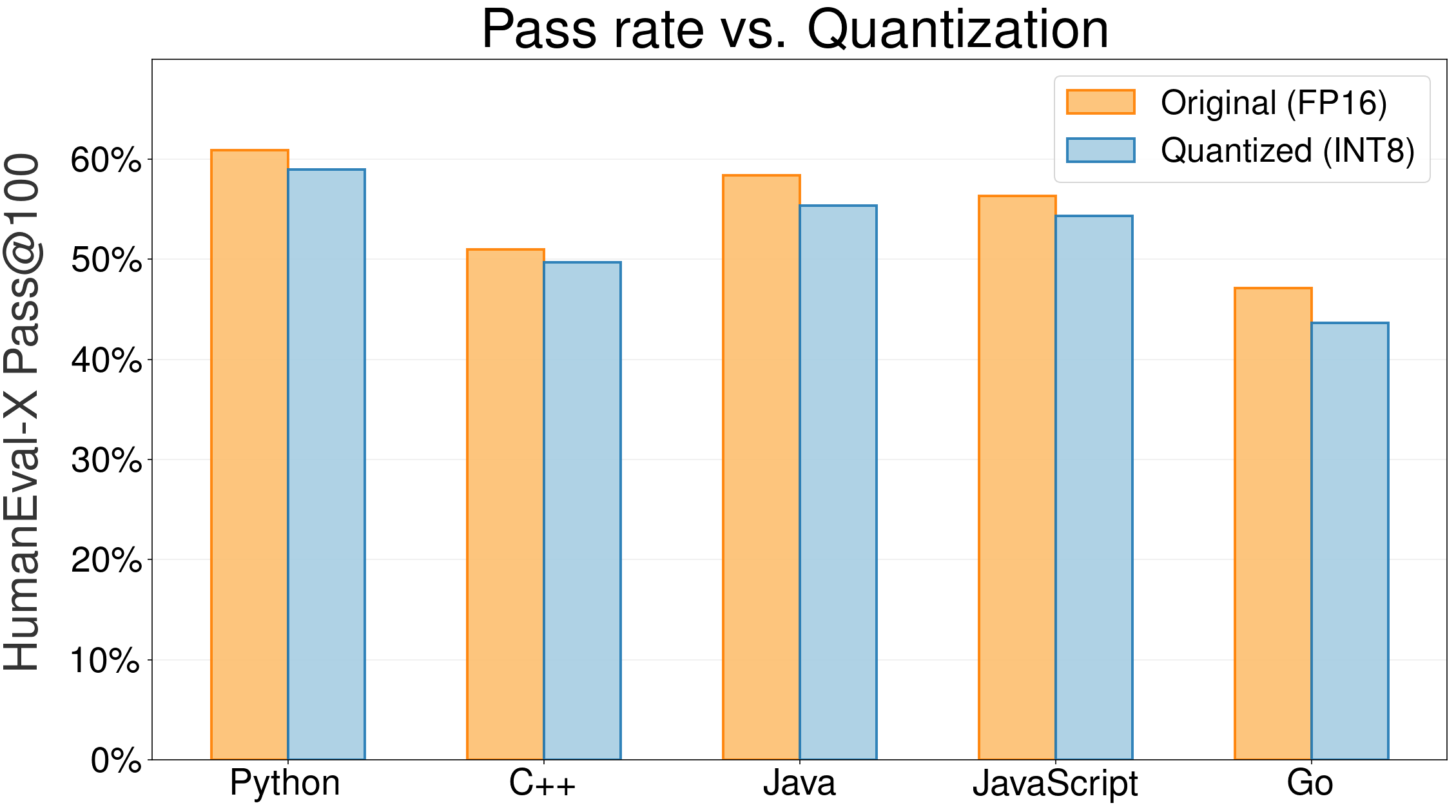}
    \caption{\name vs. its quantized version on code generation of \bench. 
    } 
    \label{fig:hx-quantize}
\end{figure}

As in~\tablename~\ref{tab:inference}, the memory consumption of \name decreases from $\sim$26.9GB to $\sim$14.7GB (down by 45.4\%), allowing CodeGeeX inference on one RTX 3090 GPU. 
Importantly, \figurename~\ref{fig:hx-quantize} shows that the quantization only slightly affects the performance on the code generation task (Cf Section~\ref{sec:bench} for details about \bench.).

\begin{table}[htbp]
\centering
\renewcommand{\arraystretch}{1.2}
\caption{GPU memory and inference time of \name w/ and w/o quantization on different GPUs and frameworks.}
\label{tab:inference}
\resizebox{\columnwidth}{!}{
\begin{tabular}{ccc|cc|cc|cc|cc|cc} 
\toprule
\multirow{2}{*}{\textbf{Implementation}} & \multirow{2}{*}{\textbf{GPU}} & \multirow{2}{*}{\textbf{Format}} & \multicolumn{2}{c}{\textbf{L=128}} & \multicolumn{2}{c}{\textbf{L=256}} & \multicolumn{2}{c}{\textbf{L=512}} & \multicolumn{2}{c}{\textbf{L=1024}} & \multicolumn{2}{c}{\textbf{L=2048}} \\
 &  &  & \textbf{Mem (G)} & \textbf{Time (s)} & \textbf{Mem (G)} & \textbf{Time (s)} & \textbf{Mem (G)} & \textbf{Time (s)} & \textbf{Mem (G)} & \textbf{Time (s)} & \textbf{Mem (G)} & \textbf{Time (s)} \\ 
\midrule
Pytorch & 3090 & FP16 & \multicolumn{10}{c}{OOM} \\
Pytorch & A100 & FP16 & 26.9 & 3.66 & 27.1 & 7.16 & 27.6 & 14.35 & 28.9 & 29.95 & 34.6 & 63.20 \\
Megatron & A100 & FP16 & 26.9 & 4.55 & 27.1 & 9.40 & 27.6 & 18.65 & 28.9 & 37.63 & 34.6 & 75.02 \\
Megatron & 2xA100 & FP16 & 17.9 & 5.11 & 22.1 & 10.17 & 22.1 & 20.42 & 22.1 & 41.04 & 22.1 & 82.93 \\
Megatron & 4xA100 & FP16 & 8.0 & 5.25 & 11.1 & 10.35 & 11.1 & 20.89 & 11.1 & 41.86 & 11.1 & 84.95 \\
Megatron & 8xA100 & FP16 & 4.8 & 5.47 & 5.7 & 11.04 & 6.3 & 22.38 & 6.5 & 45.50 & 6.5 & 90.47 \\
\midrule
Pytorch & 3090 & INT8 & 14.7 & 13.82 & 15.7 & 27.10 & 16.1 & 55.42 & 17.1 & 110.83 & 18.7 & 228.67 \\
Pytorch & A100 & INT8 & 14.7 & 9.40 & 15.7 & 18.65 & 16.1 & 37.38 & 17.1 & 75.60 & 18.7 & 155.01 \\
LLM.int8() & A100 & INT8 & 14.7 & 20.65 & 15.1 & 35.86 & 15.6 & 72.76 & 16.7 & 147.59 & 22.3 & 301.93 \\
\midrule
Oneflow & A100 & FP16 & 25.9 & 2.61 & 26.2 & 5.25 & 27.0 & 10.89 & 29.0 & 22.49 & 33.6 & 47.54 \\
Oneflow & A100 & INT8 & 13.6 & 1.85 & 13.9 & 3.73 & 14.4 & 7.83 & 15.9 & 16.24 & 21.1 & 35.98 \\
FastTrans & A100 & FP16 & 26.0 & 2.43 & 26.1 & 4.93 & 26.3 & 10.21 & 26.7 & 22.60 & 27.5 & 50.09 \\
FastTrans & A100 & INT8 & 14.9 & \textbf{1.61} & 15.0 & \textbf{3.24} & 15.2 & \textbf{6.35} & 15.6 & \textbf{14.32} & 17.4 & \textbf{34.96} \\
\bottomrule
\end{tabular}}
\end{table}

\vpara{Acceleration.} 
After quantization, we further implement a faster version of \name using the NVIDIA FasterTransformer (FastTrans). 
It supports highly-optimized operations by using layer fusion, GEMM autotuning, and hardware-accelerated functions. 
For INT8 quantized version, we also implement a custom kernel that accelerates the mixed precision matrix multiplication between INT8 weights and FP16 activation vectors. 
According to \tablename~\ref{tab:inference}, the INT8 quantization plus FastTrans implementation achieves the fastest inference speed and the lowest GPU memory consumption on a single GPU.
The inference time per token is within 13ms (1.61 seconds / 128 tokens).
We also compare the inference speed with implementations in LLM.int()~\citep{dettmers2022llm} and Oneflow~\citep{yuan2021oneflow}.
\label{sec:model}

\section{The HumanEval-X Benchmark}
We develop the \bench benchmark\footnote{The \bench dataset and docker image are at \hyperlink{https://hub.docker.com/r/codegeex/codegeex}{https://hub.docker.com/r/codegeex/codegeex}.} for evaluating multilingual code models. 
There are 164 code problems defined for five major languages: C++, Java, JavaScript, Go, and Python, resulting in 164$\times$5=820 problem-solution pairs.  
For each problem, it supports both code generation and code translation.
Examples of the problems can be found in Appendix~\ref{app:example}.

\subsection{\bench: A Multilingual Benchmark}

HumanEval~\citep{chen2021codex} has been developed to evaluate Codex by OpenAI. 
However, similar to MBPP~\citep{austin2021program} and APPS~\citep{hendrycks2021apps}, it only consists of  handcrafted  programming problems  in Python, thus cannot be directly applied to systematically evaluate the performance of multilingual code generation. 

To this end, we propose to develop a multilingual variant of HumanEval, referred to as HumanEval-X.
This is not trivial. For each problem, defined only for Python, in HumanEval, we manually rewrite its prompt, canonical solution, and test cases in the other four languages---C++, Java, JavaScript, and Go.
Altogether, we have 820 problem-solution pairs in total in \bench, each comprising the following parts:
\begin{itemize}
    \item \textbf{task\_id}: programming language and numerical problem id, \emph{e.g.}, \texttt{Java/0} represents the 0-th problem in Java; 
    \item \textbf{declaration}: function declaration including necessary libraries or packages;
    \item \textbf{docstring}: description that specifies the functionality and example input/output;
    \item \textbf{prompt}: function declaration plus docstring;
    \item \textbf{canonical\_solution}: a verified solution to the problem; 
    \item \textbf{test}: test program including test cases.
\end{itemize}

\begin{figure}[tp]
    \centering
    \includegraphics[width=\columnwidth]{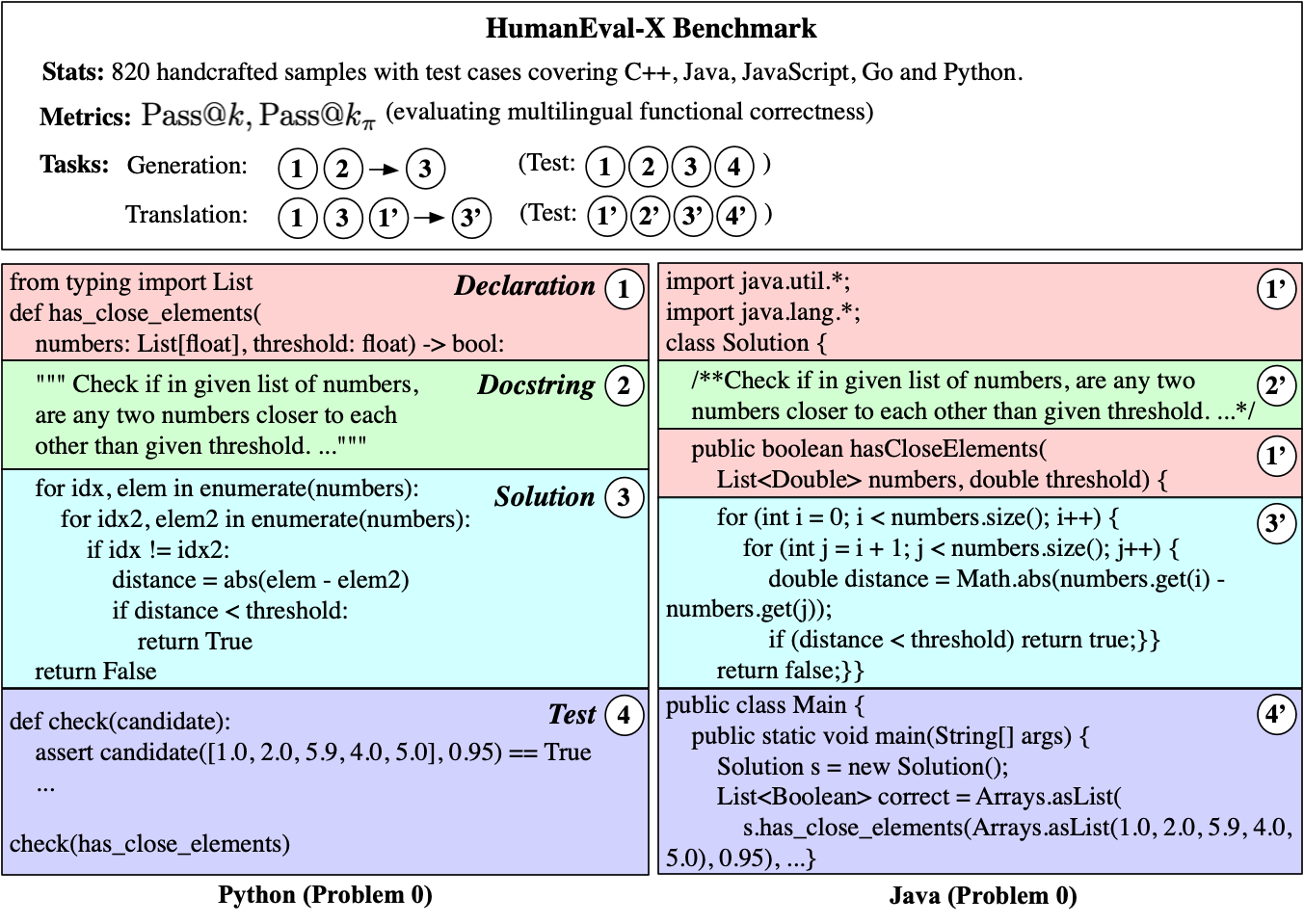}
    \vspace{-3mm}
    \caption{An illustration of code \textit{generation} and \textit{translation} tasks in \bench. 
    \textmd{Declarations, docstrings, solutions, and test cases are marked with \textcolor[RGB]{255,50,50}{red}, \textcolor[RGB]{50,232,50}{green}, \textcolor[RGB]{50,50,251}{blue}, and \textcolor[RGB]{140,15,145}{purple} respectively. \textit{Generation} uses declaration and docstring as input to generate the solution. \textit{Translation} uses declaration in both languages and solution in source language as input, to generate solution in the target language (docstring is not used to prevent models from directly solving the problem).}}
    \label{fig:hx-demo}
    \vspace{-3mm}
\end{figure}

Each problem-solution pair in \bench supports both code generation 
code translation. 
An illustrative example is shown in \figureautorefname~\ref{fig:hx-demo}.
We take the following efforts to make sure that the rewritten code conforms to the programming style of the corresponding language. 
First, we use the customary naming styles, like \texttt{CamelCase} in Java, Go, and JavaScript, and \texttt{snake\_case} in C++.
Second, we put the docstrings before the function declaration in Java, JavaScript, C++, and Go.
Symbols in docstrings are modified, \emph{e.g.}, single quotes are replaced by double quotes in some languages, and keywords like \texttt{True/False}, \texttt{None} are also replaced. 
Third, we refine test cases according to language-specific behaviors, rather than forcing the programs to return the same result for different languages.
For example, when converting an integer to a binary string, Python method \texttt{bin} adds a prefix \texttt{``0b''} before the string while Java method \texttt{Integer.toBinaryString} does not, so we remove such prefix in Java test cases.
Last, we also take care of the rounding function. In Python, \texttt{round} converts half to the closest even number, unlike in other languages.
Thus, we change the test cases to match the rounding implementations in each language.

\subsection{\bench: Tasks}

In \bench, we evaluate two tasks: code generation and code translation.

\vpara{Code Generation.}
The task of code generation takes a problem description (e.g., \texttt{``write a factorial function''}) as input and generates the solution in the selected languages (Cf Figure ~\ref{fig:codegeex_summary} (a)). 
Specifically, the model takes in the prompt including declaration and docstrings, and generates the implementation of the function. 
Note that  \bench uses the same problem set for all the five languages, thus, for solving each problem, it supports either one single language or multiple languages simultaneously.

\vpara{Code Translation.}
The task of code translation takes the implementation of a problem in the source language and generates its counterpart implementation in the target language.
Precisely, its input includes the function declaration and a canonical solution in the source language (e.g., Python).
The model should translate the  solution to the target language.
Adding declaration in the target language restricts function names and variable types, making the evaluation easier, especially under the zero-shot setting. 
To prevent the models from directly solving the problem rather than translating, we do not include the docstrings. 
\bench supports the translation between all pairs of 5 languages, that is, in total 20 source-target language pairs.

\vpara{Metric.}
For both tasks, we use test cases to evaluate the exact functional correctness of the generated code, measuring the performance with pass@$k$ \citep{kulal2019spoc}, making it real-world useful and also completely different from the string similarity metrics like BLEU~\citep{papineni2002bleu}, and CodeBLEU~\citep{ren2020codebleu,lu2021codexglue,zhu2022xlcost}. 
Specifically, we use the unbiased method to estimate pass@$k$ \citep{chen2021codex}:
\begin{equation}
\label{eqn:pass}
\small
    \text{pass}@k:= \mathbb{E}[1-\frac{\tbinom{n-c}{k}}{\tbinom{n}{k}}], n=200, k\in\{1,10,100\}
\end{equation}
where $n$ is the total number of generation ($n$=200 in this work), $k$ is the sampling budget (typically $k\in\{$1, 10, 100$\}$) and $c$ is the number of samples that pass all test cases. 
We average over the problem set to get the expectation.
$1-\frac{\tbinom{n-c}{k}}{\tbinom{n}{k}}$ is the estimated pass$@k$ for a single problem, and $\mathbb{E}$ is the expectation of pass$@k$ over all problems. In practice, we average single-problem pass$@k$ among all test-set problems to get the expectation.

\vpara{Multilingual Metric with Budget Allocation.}
Unlike mono-lingual models, multilingual code models can solve problems by allocating generation budgets to various languages to increase the sampling diversity and improve the solve rate. 
Given a budget $k$, we can distribute part of it $n_i$ to each language with the assignment 
\begin{equation}
\small
    \pi=(n_1,n_2,...,n_m), \sum_{i=1}^m n_i=k,
\end{equation}
where $n_i$ is the generation budget assigned to language $i$, $m$ is the number of candidate languages.
Under an assignment $\pi=(n_1,...n_m)$, for a problem $p$, the pass$@k_{\pi}$ can be estimated by:
\begin{equation}
\small
    \text{pass}@k_{\pi}=\mathbb{E}[1-\prod_{i=1}^m \frac{\tbinom{n-c_i}{n_i}}{\tbinom{n}{n_i}}],
\end{equation}
where $n$ is the total number of generation, $n_i$ is the sampling budget and $c_i$ is the number of samples that pass all test cases for language $i$.
We show in Section~\ref{sec:multi_ability} that multilingual models can benefit from budget allocation strategies and have higher solve rate than using any single language.

\label{sec:bench}

\section{Evaluating \name on \bench}

We evaluate \name for the code generation and translation tasks on the multilingual benchmark \bench. 
By inheriting from HumanEval, the \bench results on Python are equivalent to the evaluation on HumanEval.

\subsection{Evaluation Settings}
\label{sec:eval-setting}

\vpara{Baselines.} 
We compare \name with five competitive open-source baselines: 
GPT-J-6B~\citep{wang2021gptj}, GPT-NeoX-20B~\citep{black2022gptneox}, InCoder-6.7B~\citep{fried2022incoder}, and CodeGen-Multi-6B/16B~\citep{nijkamp2022codegen}. 
These models are all trained on multilingual code data, but is previously only evaluated in HumanEval (Python). 
And they are closer to the scale of \name or even larger, while smaller models in the literature are ignored.
For all baselines, we use the versions available on HuggingFace~\citep{wolf2019huggingface}. 
We follow  the experimental settings of \bench in Section \ref{sec:bench}. 
Further details can be found in Appendix~\ref{app:hx-additional}.

\textbf{Environment.} 
Experiments are conducted by using the NVIDIA A100-SXM-40GB GPUs with Linux system. 
We design a distributed framework for generation based on ZeroMQ to balance GPU loads. 
All generated codes are tested in language-specific environments with necessary packages installed.

\textbf{Decoding Strategy.} 
We use temperature sampling ($t\in[0, 1]$) and nucleus sampling ($p\in[0, 1]$) for generation. 
For \name in code generation, we use $t=0.2, p=0.95$ for pass@1 and $t=0.8, p=0.95$ for pass@10 and pass@100 (except for Go and JavaScript, where $p=0.9$). 
For \name in code translation, we use $t=0.2, p=0.95$ for pass@1 and $t=0.8, p=0.95$ for pass@10 and pass@100 for all language pairs. For the fine-tuned \name-13B-FT used for code translation, we use $p=0.95$. 
For all baselines in both tasks, we use $t=0.2, p=0.95$ for pass@1, $t=0.8, p=0.95$ for pass@10 and pass@100.
All pass@$k$, $k\in\{1,10,100\}$ results are estimated with $n=200$.
The maximum number of generated tokens is set to 1024 for all models.

\begin{table}
     \centering
     \caption{Results of \textbf{code generation} task in \bench. }
     \resizebox{\columnwidth}{!}{
     \begin{tabular}{cccccccc} 
     \toprule
     \textbf{Language} & \textbf{Metric} & \begin{tabular}[c]{@{}c@{}}\textbf{GPT-J}\\\textbf{-6B}\end{tabular} &
     \begin{tabular}[c]{@{}c@{}}\textbf{GPT-NeoX}\\\textbf{-20B}\end{tabular} &
     \begin{tabular}[c]{@{}c@{}}\textbf{InCoder}\\\textbf{-6.7B}\end{tabular} & \begin{tabular}[c]{@{}c@{}}\textbf{CodeGen}\\\textbf{-Multi-6B}\end{tabular} & \begin{tabular}[c]{@{}c@{}}\textbf{CodeGen}\\\textbf{-Multi-16B}\end{tabular} & \begin{tabular}[c]{@{}c@{}}\textbf{CodeGeeX}\\\textbf{-13B (ours)}\end{tabular} \\
     \midrule
     \multirow{3}{*}{\textbf{Python}} & pass@1 & 11.10\% & 13.83\% & 16.41\% & 19.41\% & 19.22\% & \textbf{22.89\%} \\
     & pass@10 & 18.67\% & 22.72\% & 26.55\% & 30.29\% & 34.64\% & \textbf{39.57\%} \\
     & pass@100 & 30.98\% & 39.56\% & 43.95\% & 49.63\% & 55.17\% & \textbf{60.92\%} \\
     \midrule
     \multirow{3}{*}{\textbf{C++}} & pass@1 & \ 7.54\% & \ 9.90\% & \ 9.50\% & 11.44\% & \textbf{18.05\%} & 17.06\% \\
     & pass@10 & 13.67\% & 18.99\% & 19.30\% & 26.23\% & 30.84\% & \textbf{32.21\%} \\
     & pass@100 & 30.16\% & 38.75\% & 36.10\% & 42.82\% & 50.90\% & \textbf{51.00\%} \\
     \midrule
     \multirow{3}{*}{\textbf{Java}} & pass@1 & \ 7.86\% & \ 8.87\% & \ 9.05\% & 15.17\% & 14.95\% & \textbf{20.04\%} \\
     & pass@10 & 14.37\% & 19.55\% & 18.64\% & 31.74\% & \textbf{36.73\%} & 36.70\% \\
     & pass@100 & 32.96\% & 42.23\% & 40.70\% & 53.91\% & \textbf{60.62\%} & 58.42\% \\
     \midrule
     \multirow{3}{*}{\textbf{JavaScript}} & pass@1 & \ 8.99\% & 11.28\% & 12.98\% & 15.41\% & \textbf{18.40\%} & 17.59\% \\
     & pass@10 & 16.32\% & 20.78\% & 22.98\% & 27.92\% & \textbf{32.80\%} & 32.28\% \\
     & pass@100 & 33.77\% & 42.67\% & 43.34\% & 48.81\% & \textbf{56.48\%} & 56.33\% \\
     \midrule
     \multirow{3}{*}{\textbf{Go}} & pass@1 & \ 4.01\% & \ 5.00\% & \ 8.68\% & \ 9.98\% & 13.03\% & \textbf{14.43\%} \\
     & pass@10 & 10.81\% & 15.70\% & 13.80\% & 23.26\% & 25.46\% & \textbf{25.68\%} \\
     & pass@100 & 23.70\% & 32.08\% & 28.31\% & 41.01\% & \textbf{48.77\%} & 47.14\% \\
     \midrule
     \multirow{3}{*}{\textbf{Average}} & pass@1 & \ 7.90\% & \ 9.78\% & 11.33\% & 14.28\% & 16.73\% & \textbf{18.40\%} \\
     & pass@10 & 14.77\% & 19.55\% & 20.25\% & 27.89\% & 32.09\% & \textbf{33.29\%} \\
     & pass@100 & 30.32\% & 39.06\% & 38.48\% & 47.24\% & 54.39\% & \textbf{54.76\%} \\
     \bottomrule
     \end{tabular}}
     \label{tab:sing}
     \vspace{-3mm}
\end{table}

\begin{figure}
    \centering
    \includegraphics[width=\textwidth]{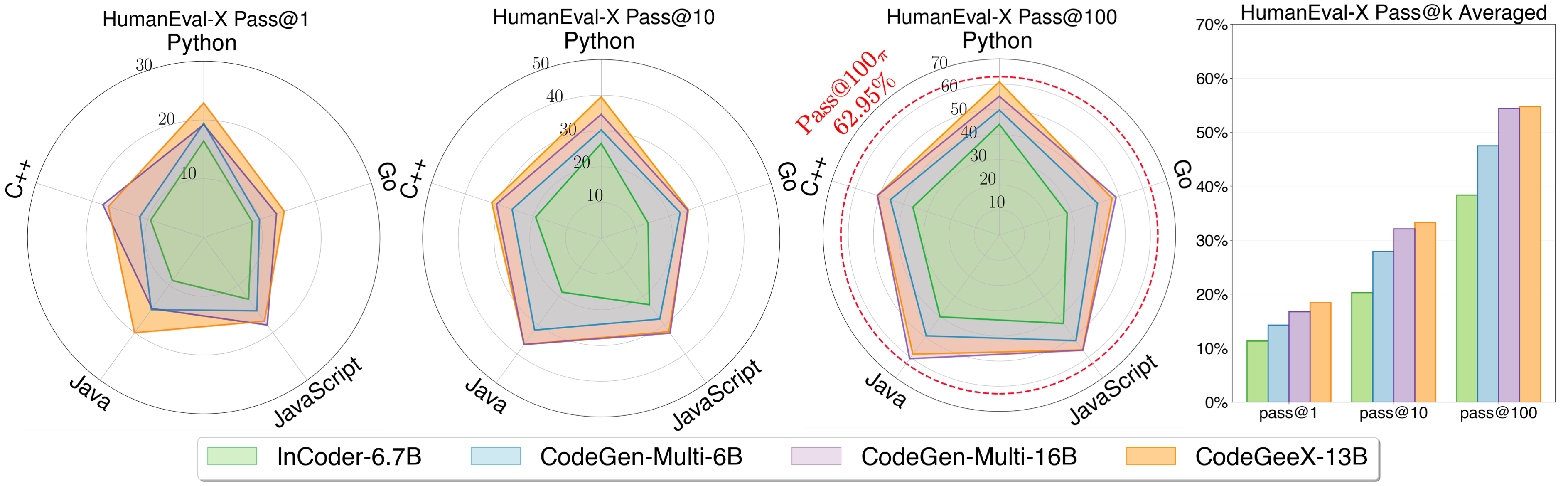}
    \caption{Results of \textbf{code generation} task in \bench. 
    {Left}: Detailed pass@k performance in five languages. 
    {Right}: \name achieves the highest average performance compared with other open-sourced multilingual baselines. We also find that it gains performance when the sampling budgets are properly distributed to multiple languages.} 
    \label{fig:hx-radar}
\end{figure}

\subsection{Results of Code Generation and Translation}

\textbf{Multilingual Code Generation.}
\tablename~\ref{tab:sing} and \figurename~\ref{fig:hx-radar} report the code generation results in terms of the pass@$k$, $k\in\{1,10,100\}$ for \name and five baseline models on five programming languages. 
\name significantly outperforms models trained with mixed corpora (GPT-J-6B and GPT-NeoX-20B), even though GPT-NeoX-20B has much more parameters. 
For models trained on codes, \name outperforms those with smaller scales (InCoder-6.7B, CodeGen-Multi-6B) by a large margin, and is competitive with CodeGen-Multi-16B with a larger scale.
\name achieves the best average performance among all models, even slightly better than the larger CodeGen-Multi-16B in all three metrics (0.37\%$\sim$1.67\% improvements).
When considering individual languages, models have preferences highly related to the training set distribution.
For example, the best language for \name is Python while the best language for CodeGen-Multi-16B is Java.
Examples of \name generation can be found in~\appendixautorefname~\ref{app:example}.

\begin{table}

        \renewcommand{\arraystretch}{1.3}
        \centering
    \caption{Results of \textbf{code translation} task in \bench.}
    \resizebox{\columnwidth}{!}{
    \begin{tabular}{c|c|ccccccccccccccc} 
    \toprule
     &  & \multicolumn{15}{c}{\textbf{Target Language}} \\
     & \multirow{2}{*}{\textbf{Model}} & \multicolumn{3}{c}{\textbf{Python}} & \multicolumn{3}{c}{\textbf{C++}} & \multicolumn{3}{c}{\textbf{Java}} & \multicolumn{3}{c}{\textbf{JavaScript}} & \multicolumn{3}{c}{\textbf{Go}} \\
     &  & @1 & @10 & @100 & \multicolumn{1}{c}{@1} & \multicolumn{1}{c}{@10} & \multicolumn{1}{c}{@100} & \multicolumn{1}{c}{@1} & \multicolumn{1}{c}{@10} & \multicolumn{1}{c}{@100} & \multicolumn{1}{c}{@1} & \multicolumn{1}{c}{@10} & \multicolumn{1}{c}{@100} & \multicolumn{1}{c}{@1} & \multicolumn{1}{c}{@10} & \multicolumn{1}{c}{@100} \\
     \midrule\multirow{4}{*}{\textbf{Py}} & InCoder-6.7B & - & - & - & 26.11 & 41.00 & 54.25 & 26.74 & 42.66 & 61.20 & 37.05 & 58.85 & 78.91 & 15.69 & 27.57 & 43.67 \\
     & CodeGen-Multi-16B & - & - & - & \textbf{35.94} & \textbf{47.81} & 59.37 & 29.27 & 45.70 & 64.45 & \textbf{43.40} & \textbf{66.26} & \textbf{82.55} & \textbf{28.87} & \textbf{41.01} & \textbf{57.72} \\
     & CodeGeeX-13B & - & - & - & 26.54 & 43.56 & 56.48 & 25.84 & 41.52 & 59.72 & 23.22 & 47.33 & 65.87 & \ 9.56 & 23.83 & 33.56 \\
     & CodeGeeX-13B-FT & - & - & - & 34.16 & 46.86 & \textbf{61.22} & \textbf{41.98} & \textbf{58.17} & \textbf{72.78} & 34.81 & 53.05 & 66.08 & 16.41 & 30.76 & 46.37 \\
     \midrule
    \multirow{4}{*}{\textbf{C++}} & InCoder-6.7B & 34.37 & 58.41 & 78.57 & - & - & - & 34.04 & 57.02 & 68.70 & 37.05 & 65.05 & 79.61 & 25.54 & 39.11 & 58.02 \\
     & CodeGen-Multi-16B & 33.83 & 55.37 & 76.64 & - & - & - & 43.20 & 69.84 & \textbf{88.82} & \textbf{54.51} & \textbf{71.50} & \textbf{83.14} & \textbf{27.94} & \textbf{49.73} & \textbf{68.32} \\
     & CodeGeeX-13B & 27.18 & 49.02 & 67.69 & - & - & - & 22.56 & 40.91 & 64.08 & 30.23 & 55.68 & 75.58 & \ 8.64 & 18.79 & 31.76 \\
     & CodeGeeX-13B-FT & \textbf{62.79} & \textbf{80.39} & \textbf{87.10} & - & - & - & \textbf{71.68} & \textbf{81.62} & 85.84 & 50.83 & 64.55 & 74.57 & 16.71 & 34.18 & 52.98 \\
     \midrule
    \multirow{4}{*}{\textbf{Java}} & InCoder-6.7B & 42.76 & 65.55 & 80.43 & 40.01 & 55.17 & 70.39 & - & - & - & 43.20 & \textbf{68.24} & \textbf{84.39} & 21.58 & 35.20 & 54.97 \\
     & CodeGen-Multi-16B & 52.73 & 69.30 & 82.74 & 41.42 & 54.68 & 65.50 & - & - & - & \textbf{57.65} & 67.90 & 79.22 & \textbf{34.00} & \textbf{48.49} & \textbf{67.94} \\
     & CodeGeeX-13B & 43.41 & 68.46 & 84.03 & 39.33 & 58.48 & 72.36 & - & - & - & 44.19 & 64.22 & 82.89 & 17.17 & 32.74 & 47.71 \\
     & CodeGeeX-13B-FT & \textbf{75.03} & \textbf{87.71} & \textbf{95.13} & \textbf{49.67} & \textbf{65.65} & \textbf{75.40} & - & - & - & 49.95 & 62.82 & 79.64 & 18.85 & 32.92 & 48.93 \\
     \midrule   
    \multirow{4}{*}{\textbf{JS}} & InCoder-6.7B & 23.18 & 50.47 & 67.26 & 35.47 & 54.48 & 70.71 & 30.67 & 50.90 & 71.03 & - & - & - & 25.79 & 42.96 & 61.47 \\
     & CodeGen-Multi-16B & 35.52 & 52.23 & 69.78 & 35.41 & 53.12 & 64.47 & 33.79 & 56.06 & 74.00 & - & - & - & \textbf{33.38} & \textbf{49.08} & \textbf{64.14} \\
     & CodeGeeX-13B & 31.15 & 54.02 & 72.36 & 30.32 & 51.63 & 69.37 & 24.68 & 48.35 & 69.03 & - & - & - & 11.91 & 26.39 & 39.81 \\
     & CodeGeeX-13B-FT & \textbf{67.63} & \textbf{81.88} & \textbf{89.30} & \textbf{46.87} & \textbf{60.82} & \textbf{73.18} & \textbf{56.55} & \textbf{70.27} & \textbf{80.71} & - & - & - & 16.46 & 32.99 & 50.29 \\
     \midrule
    \multirow{4}{*}{\textbf{Go}} & InCoder-6.7B & 34.14 & 54.52 & 70.88 & 30.45 & 48.47 & 62.81 & 34.52 & 53.95 & 69.92 & 39.37 & \textbf{63.63} & \textbf{80.75} & - & - & - \\
     & CodeGen-Multi-16B & 38.32 & 50.57 & 68.65 & 32.95 & 45.88 & 59.56 & 36.55 & 59.12 & 78.70 & 38.93 & 56.68 & 70.68 & - & - & - \\
     & CodeGeeX-13B & 35.92 & 56.02 & 77.32 & 29.83 & 41.98 & 58.15 & 22.89 & 41.04 & 61.46 & 25.24 & 46.50 & 69.93 & - & - & - \\
     & CodeGeeX-13B-FT & \textbf{57.98} & \textbf{79.04} & \textbf{93.57} & \textbf{38.97} & \textbf{53.05} & \textbf{63.92} & \textbf{54.22} & \textbf{69.03} & \textbf{79.40} & \textbf{43.07} & 59.78 & 74.04 & - & - & - \\
    \bottomrule
    \end{tabular}}
    \label{tab:trans}
\end{table}

\textbf{Cross-Lingual Code Translation.}
\tablename~\ref{tab:trans} illustrates the results on code translation.
For \name, we evaluate both the original version \name-13B and the fine-tuned \name-13B-FT.
\name-13B-FT is first fine-tuned using the training set of code translation task in XLCoST~\citep{zhu2022xlcost}, and then continuously fine-tuned by a small amount of Go data (since Go is missing in XLCoST).
Among all translation pairs, \name-13B-FT performs the best on pass@100 in 11 out of the 20, while CodeGen-Multi-16B is the best on 7 of them.
We also observe a clear preference of languages by different models. 
\name performs the best when translating other languages to Python and C++, while CodeGen-Multi-16B performs better when translating to JavaScript and Go. 

\textbf{Test Result Analysis.}
We group the samples' test results into five categories: passing, wrong answer, runtime error, syntax/semantic error and unfinished generation, and calculate the proportion of results for each model. Runtime error includes out-of-bound index, wrong string format, etc; syntax/semantic error indicates errors detected by syntax or semantic check, like compilation error in compiled languages and syntax, undefined or type error in interpreted languages; unfinished generation means failing to complete one function within maximum length. 

\begin{figure}[htp]
\centering
\begin{subfigure}{.513\textwidth}
    \centering
    \includegraphics[width=\textwidth]{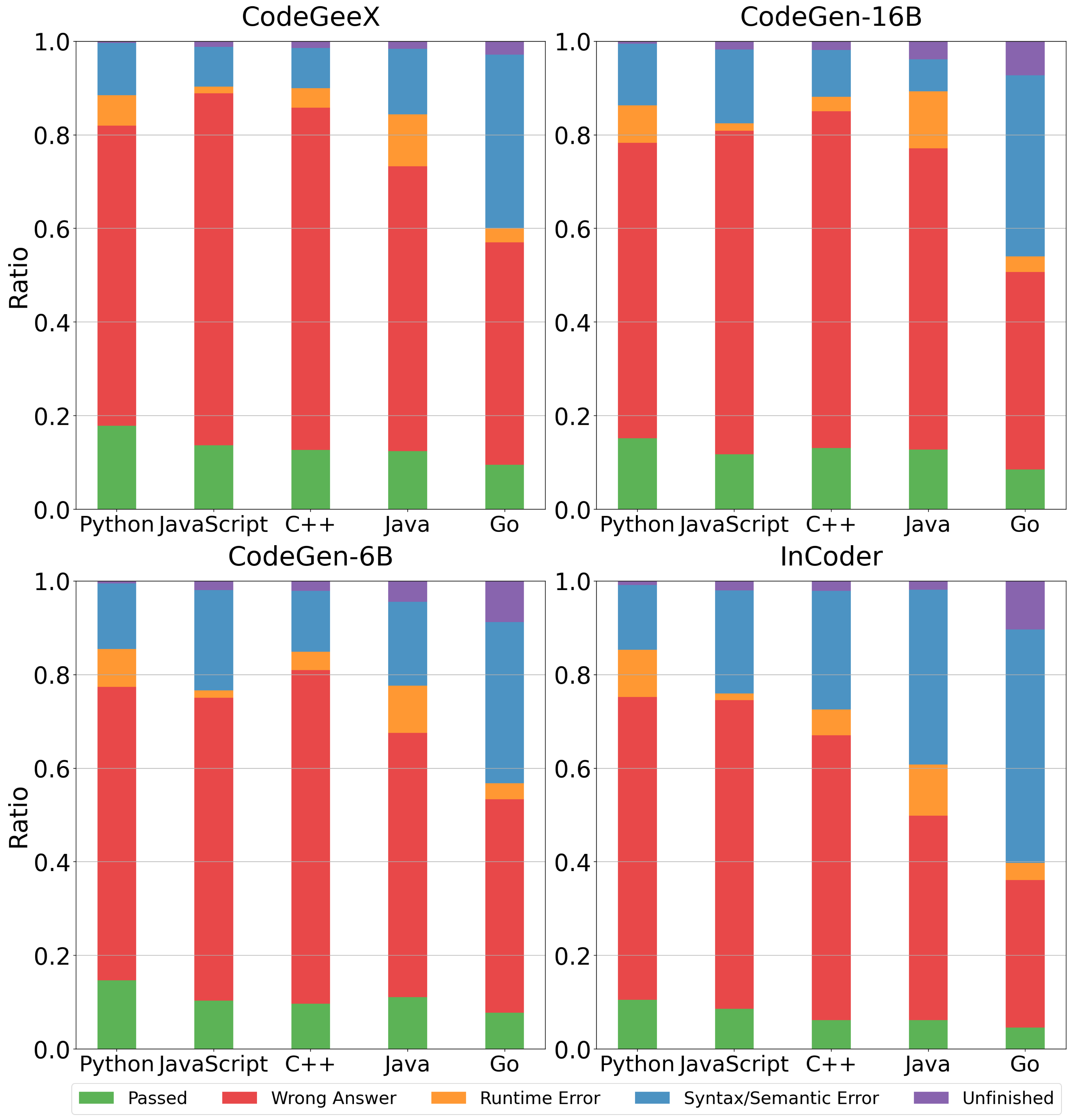}
\end{subfigure}
\begin{subfigure}{.378\textwidth}
    \centering
    \includegraphics[width=\textwidth]{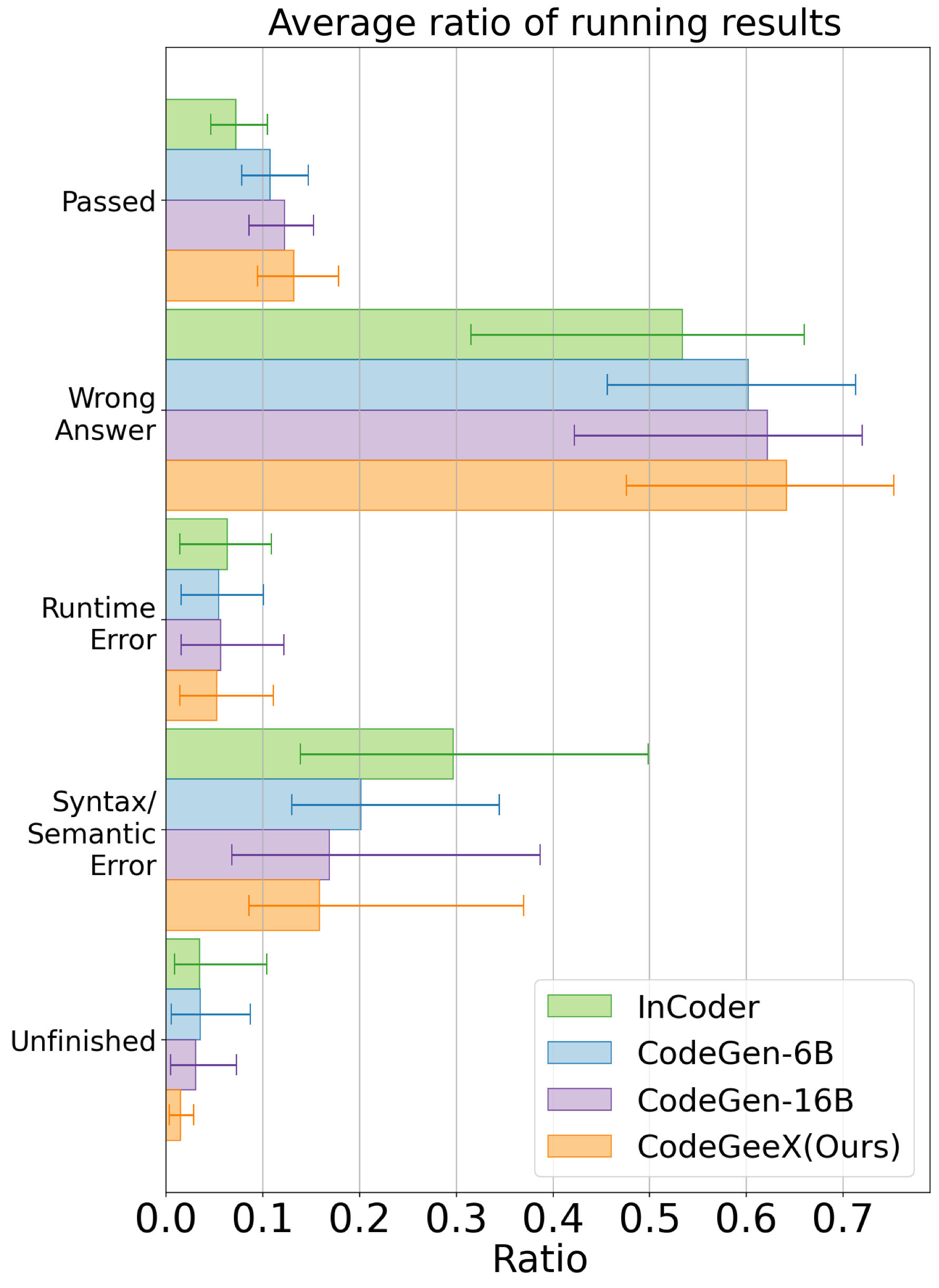}
\end{subfigure}
\caption{\textbf{Left}: the proportions of running results of four models for each language. \textbf{Right}: the average result ratios across four models, with lines representing minimum and maximum values. For each model and each language, we study 200 samples generated under $t=0.8$ and $p=0.95$.}
\label{fig:errors}
\end{figure}

\figureautorefname~\ref{fig:errors} shows the proportions of running results of four models. 
For all languages, the most common error type is wrong answer, with ratio ranging from 0.44 to 0.75 except for Go, showing that code generation models at the current stage mainly suffer from incorrect code logic rather than semantics. 
Go samples have a high syntax error rate, which may be due to Go having strict restrictions on syntax and forbidding unused variables and imports, failing to compile many logically correct codes.
\name has less rate to generate code that produces runtime, syntax, or semantic errors.
\subsection{The Multilingual Pre-Training Helps Problem Solving}
\label{sec:multi_ability}

\begin{figure}
    \centering
    \begin{minipage}[t]{0.48\textwidth}
    \centering
    \includegraphics[width=\textwidth]{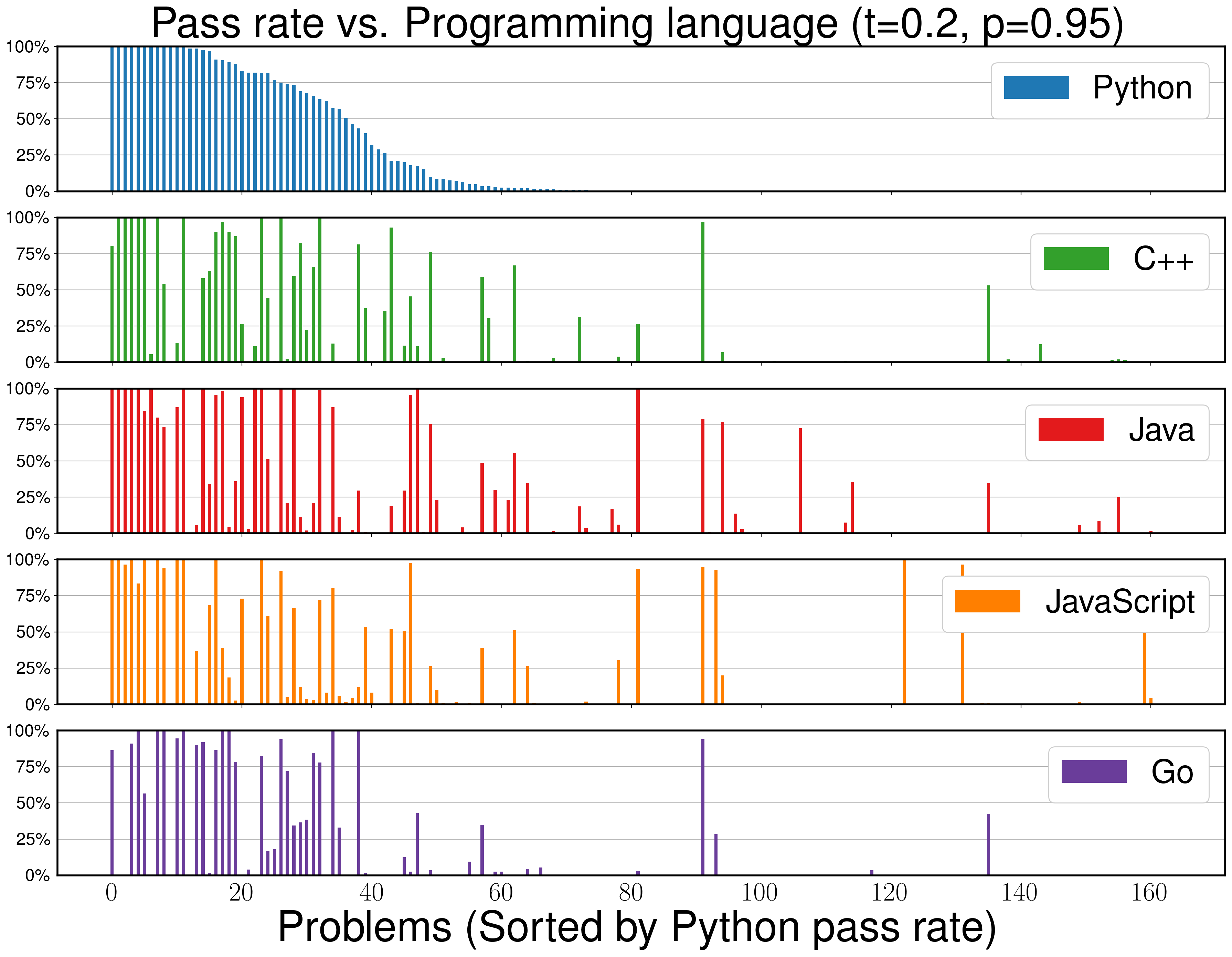}
    \end{minipage}
    \begin{minipage}[t]{0.48\textwidth}
    \centering
    \includegraphics[width=\textwidth]{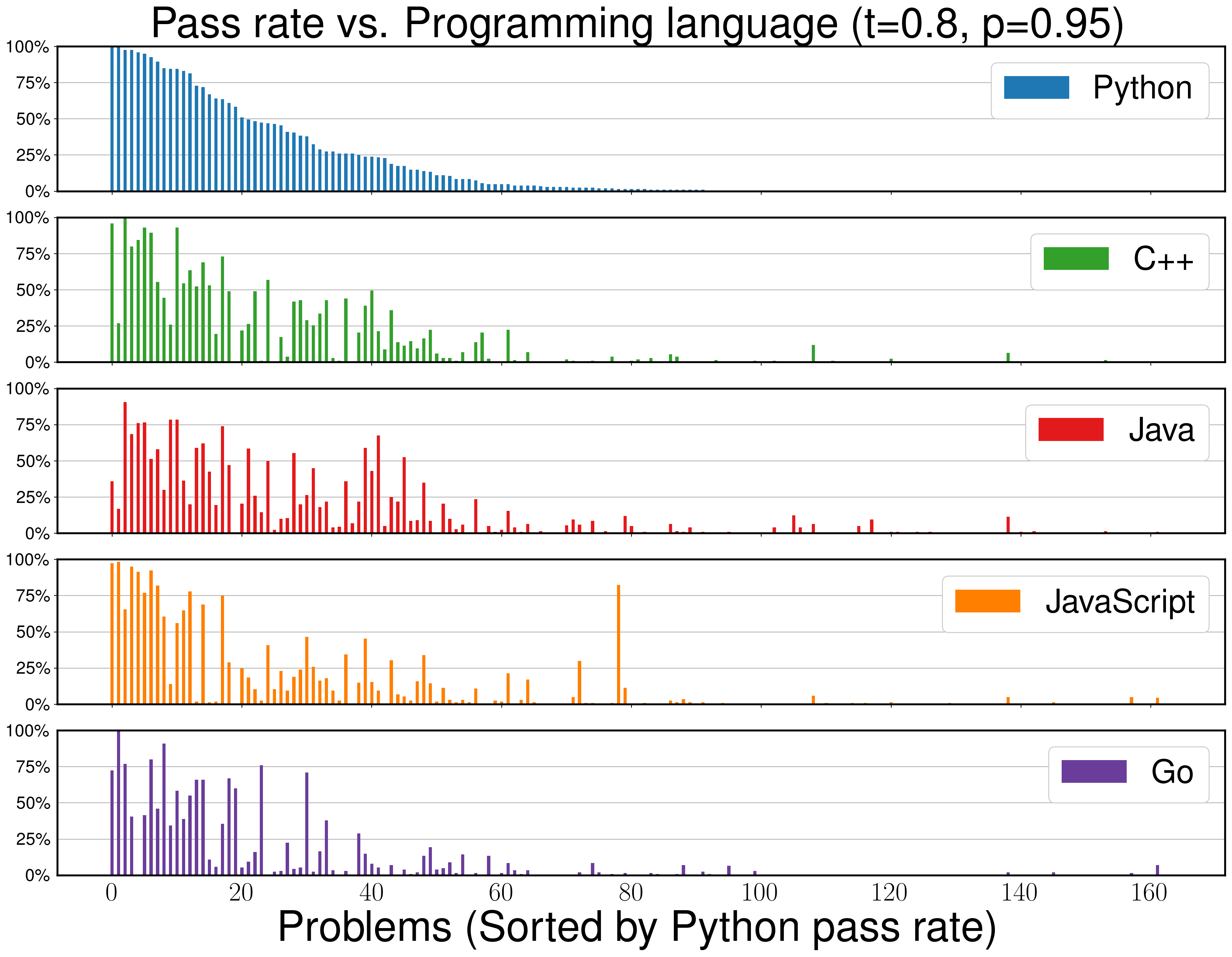}
    \end{minipage}
    \caption{In \bench, each problem's pass rate varies when generating in different programming languages with \name. \textbf{Left:} $t=0.2, p=0.95$; \textbf{Right:} $t=0.8, p=0.95$.}
    \label{fig:hx-distribution}
\end{figure}

We perform studies to understand whether and how multilingual pre-training can benefit  problem-solving  of \name.

\textbf{Exploration vs. Exploitation under Fixed Budgets.}
Given a fixed budget $k$, pass@k evaluates the ability of models generating at least 1 correct solution under $k$ generations.
Previous works~\citep{chen2021codex, li2022alphacode} have already discovered that there's a trade-off between exploration and exploitation: When the budget is small, it is better to use a low temperature to ensure accuracy on easy problems. 
When the budget is large, instead, adjusting a higher temperature is vital, as it makes the model more likely to find at least one solution for difficult problems.

\textbf{Pass Rate Distribution vs. Languages.}
Unlike monolingual models, multilingual models can solve problems using various programming languages.
In~\figurename~\ref{fig:hx-distribution}, we observe that the pass rate distribution of problems against different languages are diverse.
This inspires us to use budget allocation methods to help improve the diversity of the generated solutions.

\textbf{Budget Allocation Strategies.}
We compare three basic strategies: \textit{Best Single} chooses a single language with the best performance; \textit{Uniform} allocates the budget uniformly; \textit{Weighted} allocates the budget to multiple languages based on their proportions in the training corpus (detailed weights can be found in Appendix~\tablename~\ref{tab:app_alloc}). 
\tablename~\ref{tab:budget_alloc} illustrates how budget allocation improves the multilingual generation.
Both \textbf{Uniform} and \textbf{Weighted} outperform \textbf{Best Single} by promoting a more diverse generation, which gives a higher chance of solving problems.
\textbf{Weighted} is slightly better due to the prior knowledge on the model. 
For model-wise comparison, \name shows up a decent advantage over other baselines in both strategies, which suggests that it might have a more diverse solution set under multiple languages.
Programming languages are created with a specific purpose and unique design; in real-world scenarios, multilingual models might take this advantage for certain tasks.

\begin{table}
     \centering
     \caption{Results for fixed-budget multilingual generation on \bench. 
     }
     \resizebox{\columnwidth}{!}{
     \begin{tabular}{c|ccccccc} 
     \toprule
    \textbf{Metric} &\textbf{Method} &  \begin{tabular}[c]{@{}c@{}}\textbf{GPT-J}\\\textbf{-6B}\end{tabular} &
     \begin{tabular}[c]{@{}c@{}}\textbf{GPT-NeoX}\\\textbf{-20B}\end{tabular} &\begin{tabular}[c]{@{}c@{}}\textbf{InCoder}\\\textbf{-6.7B}\end{tabular} & \begin{tabular}[c]{@{}c@{}}\textbf{CodeGen}\\\textbf{-Multi-6B}\end{tabular} & \begin{tabular}[c]{@{}c@{}}\textbf{CodeGen}\\\textbf{-Multi-16B}\end{tabular} & \begin{tabular}[c]{@{}c@{}}\textbf{CodeGeeX}\\\textbf{-13B}\end{tabular} \\
    \midrule
      \multirow{3}*{\begin{tabular}[c]{@{}c@{}}$\text{pass}@k_{\pi}$\\($k=100$)\end{tabular}} & Best Single &33.77\%&42.67\%& 43.95\% & 53.19\% & 60.62\% & \textbf{60.92\%} \\
     & Uniform &36.40\%&44.75\%& 43.89\% & 53.47\% & 61.01\% & \textbf{62.41\%} \\
     & Weighted &\textit{36.76\%}&\textit{44.97\%}&\textit{45.60\%}&\textit{53.94\%}&\textit{61.34\%}&\textit{\textbf{62.95\%}}\\
    \bottomrule
    \end{tabular}}
    \label{tab:budget_alloc}
\end{table}

\begin{figure}[htp]
    \centering
    \includegraphics[width=\textwidth]{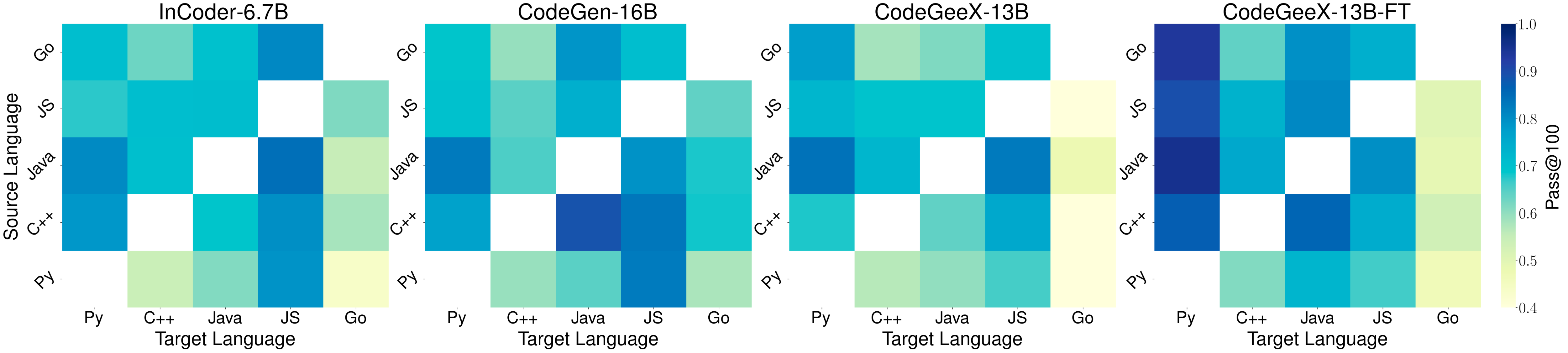}
    \caption{The performance of translating A-to-B is negatively correlated with B-to-A. Such asymmetry indicates that multilingual models still lack of high-level understanding between languages.} 
    \label{fig:analysis-translation}
\end{figure}

\textbf{Negative Correlations in Pair-Language Translation.}
When evaluating the translation ability in \bench, an interesting observation is that the performance of A-to-B and B-to-A are usually negatively-correlated, shown in~\figurename~\ref{fig:analysis-translation}.
Such asymmetry suggests that multilingual code generation models may have imbalanced focus on source and target languages during code translation. We provide two possible explanations.
First, language distributions in training corpus differ a lot, resulting in different level of generation ability.
For example, the ratio of Python is 26.6\% (vs. Go 4.7\%) in CodeGeeX training corpus, and average pass@100 of \textit{Others-to-Python} reaches \textasciitilde{}90\% (vs. \textit{Others-to-Go} only \textasciitilde{}50\%).
Second, some languages are themselves harder to automatically write with syntactic and semantic accuracy due to language-dependent features, affecting translation performance as target languages.
For instance, Go, which models translate poorly into, has more constraints on syntax level, forbidding unused variables or imports.

\section{The \name Tools and Users}
\label{sec:userexp}
 
Based on \name, we build open-source extensions for IDEs including VS Code, JetBrains and Cloud Studio.
The extensions support code generation, completion, translation and explanation, aiming at improving the development efficiency of programmers. 
As of this writing, \name has served tens of thousands of users, with an average of 250+ API calls per active user per weekday.
It currently generates 4.7+ billion tokens per week, which has been steadily growing since its release. 

\begin{figure}[htbp]
    \centering
    \begin{minipage}{0.45\textwidth}
    \includegraphics[width=\textwidth]{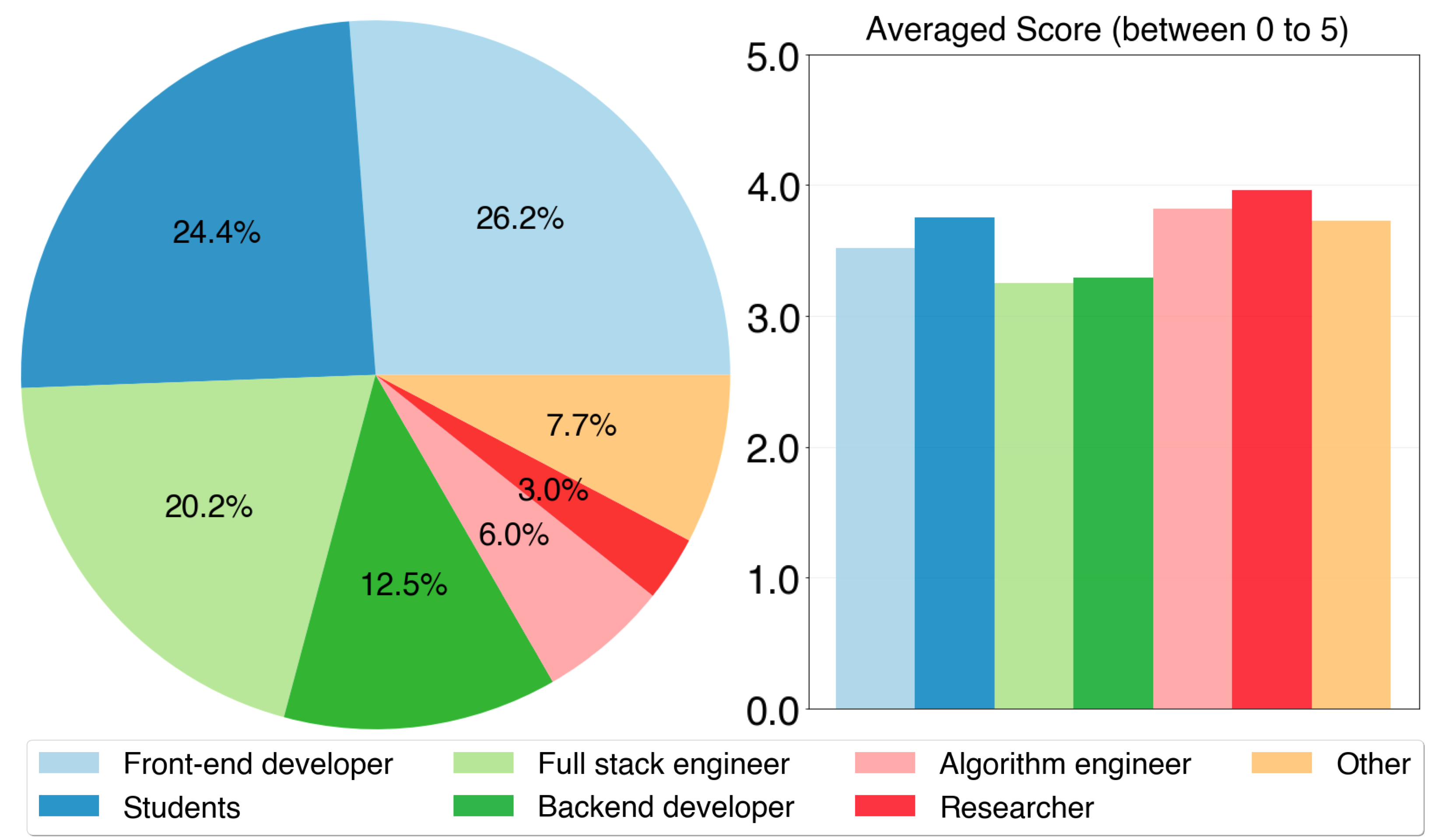}
    \captionof{figure}{Profession vs. satisfaction. Left: \textmd{Profession distribution.} Right: \textmd{Averaged rating score of \name extensions.}}
    \label{fig:survey-prof}
    \end{minipage}\hspace{2mm}
    \begin{minipage}{0.50\textwidth}
    \includegraphics[width=\textwidth]{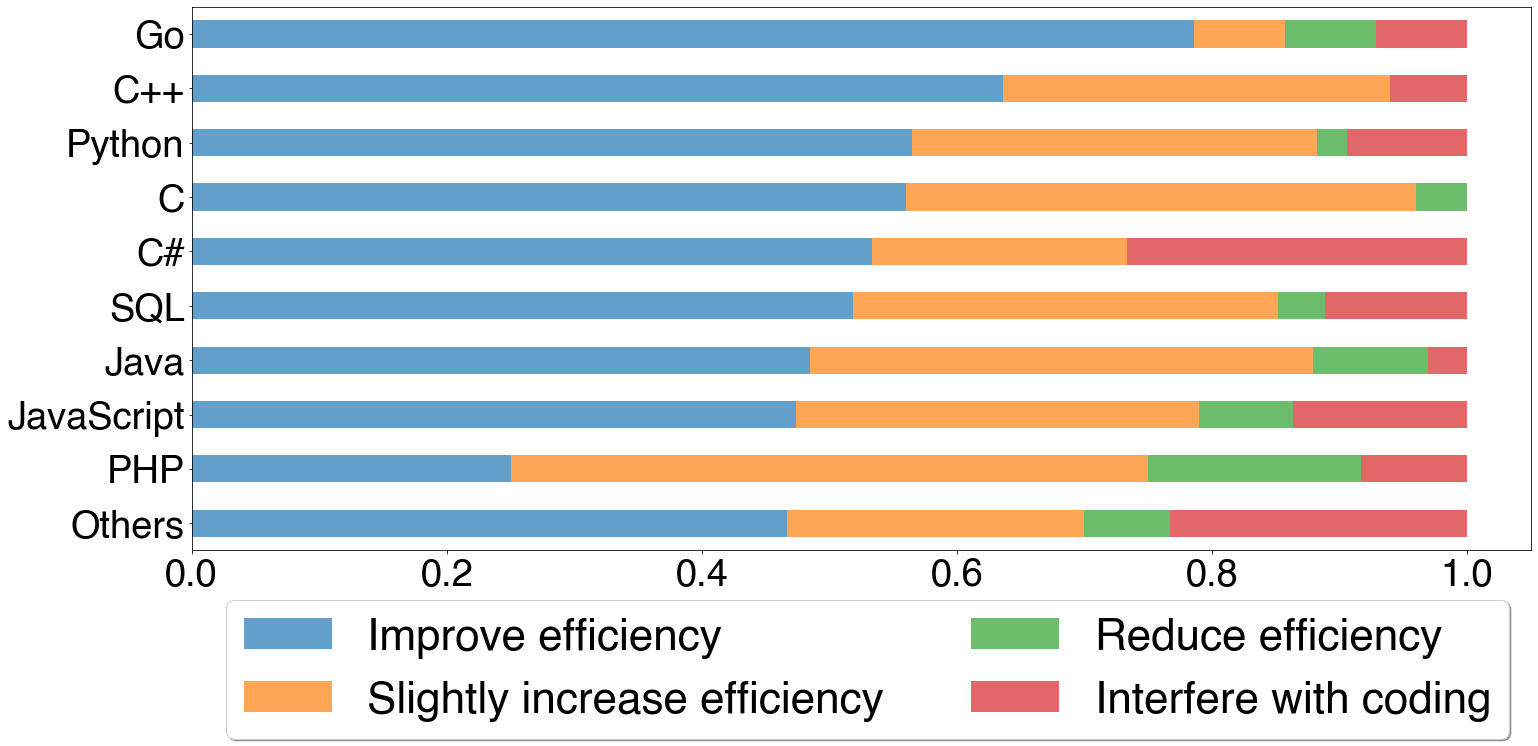}
    \captionof{figure}{Survey on "Has \name improved your coding efficiency?". \textmd{Over 83.4\% of users have positive answers.}}
    \label{fig:survey-efficiency-lang}
    \end{minipage}
\end{figure}

We perform a survey on \name's user experience from 168 users covering \textit{front-end developer}, \textit{backend developer}, \textit{full stack engineer}, \textit{algorithm engineer}, \textit{students}, \textit{researcher}, and \textit{other programmers}.
\figurename~\ref{fig:survey-prof} illustrates users' profession distribution and the satisfaction score. 
We evaluate the satisfaction considering five dimensions, "Ease of Use", "Reliability", "Feature", "Visual", "Speed", each scored from 0 to 5. 
\figurename~\ref{fig:survey-prof} shows that the majority of users have positive experiences with \name, especially for researchers and students, while there is still room for improvement for professional developers.
This can be interpreted by our training code corpus: open-sourced repositories contain many introductory or research projects, while production codes are often close-sourced. 
To increase the \name's capability in professional domain, these codes are needed in the future.

We further investigate how multilinguality of \name help coding. 
\figurename~\ref{fig:survey-efficiency-lang} illustrates how users evaluate the helpfulness of \name during development.
There are on average over 83.4\% of users think \name can improve or slightly increase their coding efficiency, especially for mainstream programming languages like Go, C++, Python, C, C\#, etc. 
Note that these well-performing programming languages also appear more frequently in the training data (\figurename~\ref{fig:data_pie}), which encourages us to train \name on more language-specific data to enhance its capability.

\section{Conclusion}

We introduce \name, a 13B pre-trained 23-language code generation model, as well as we build \bench, to fill the gap of multilingual code generation. 
\name consistently outperforms open-sourced multilingual baselines of the same scale on code generation and translation tasks.
The extensions built on \name bring significant benefits in increasing coding efficiency. 
The multilinguality of \name brings the potential of solving problems with an ubiquitous set of formalized languages. 
We open sourced \name aiming to help researchers and developers to widely take benefit of large pre-trained models for code generation.

The multilingual ability of \name shows the potential of solving problems with a ubiquitous set of formalized languages. 
Here, we share three of our observations as the future directions.

First, we find that the model capacity is essential for multilingual programming ability. 
It is not trivial for the model to benefit from learning multiple languages. 
Human programmers can abstract the high-level concept of programming, thus learning one language can help them master the others. 
On the contrary, the model seems to require a large capacity to concurrently store the knowledge of each language. 
How to help the model extract the most essential knowledge of programming remains a research challenge. 

Second, similar to others, \name shows the reasoning potential as a model though its lack of strong generality. 
We demonstrate that \name can  solve problems in different languages. 
However, the pass rate distribution varies a lot among languages, \emph{i.e.}, it is not able to solve the same problem using different languages on occasion. 
We assume that this could probably be related to some language-specific features (\emph{e.g.}, some problems are easier to solve in Python), or it could be simply due to the appearance of a similar language-specific implementation in training data. Either case, there is a long way to go for the model to have a reliable reasoning ability.

Third, the few-shot ability of \name is worth exploration. 
Instead of using costly fine-tuning approaches, we may do priming using a few examples and make the model achieve comparable performance. 
Recent works like chain-of-thought (CoT) prompting~\citep{wei2022chain} have shown impressive results using such an approach, inspiring us to examine CoT in code models.

\section*{Acknowledgement}
This research was supported by Natural Science Foundation of China (NSFC) for Distinguished Young Scholars No. 61825602, NSFC No. 62276148 and a research fund from Zhipu.AI. 
We give our special thanks to Wenguang Chen from Tsinghua, the Peng Cheng Laboratory, and Zhipu.AI for sponsoring the training and inference GPU resources. 
We thank all our collaborators and partners from Tsinghua KEG, IIIS, Peng Cheng Laboratory, and Zhipu.AI, 
including 
Aohan Zeng, Wendi Zheng, Lilong Xue, 
Yifeng Liu, Yanru Chen, Yichen Xu, 
Qingyu Chen, Zhongqi Li, Gaojun Fan,
Yifan Yao, Qihui Deng, Bin Zhou, 
Ruijie Cheng, Peinan Yu, Jingyao Zhang, Bowen Huang, Zhaoyu Wang, 
Jiecai Shan, Xuyang Ding, Xuan Xue, 
and 
Peng Zhang.

\bibliography{reference}
\bibliographystyle{acl_natbib}

\newpage
\appendix
\section{Appendix}\label{sec:appendix}
\tableofcontents
\newpage

\subsection{Statistics of Code Corpus}

\tablename~\ref{tab:dataset-comp} summarizes the composition of \name's code corpus.

\begin{table}[htbp]
    \centering
    \caption{Composition of our code corpus for pre-training.}
    \resizebox{.7\columnwidth}{!}{
    \setlength{\tabcolsep}{2mm}{
    \begin{tabular}{cccc}
    \toprule
    \textbf{Language} & \textbf{\# Tokens~(B)} & \textbf{\% Tokens~(\%)} & \textbf{Language Tag} \\
    \hline
    \textbf{C++} & 45.2283 & 28.4963 & // language: C++ \\
    \textbf{Python} & 42.3250 & 26.667 & \# language: Python \\
    \textbf{Java} & 25.3667 & 15.9824 & // language: Java \\
    \textbf{JavaScript} & 11.3165 & 7.13 & // language: JavaScript \\
    \textbf{C} & 10.6590 & 6.7157 & // language: C \\
    \textbf{Go} & 7.4774 & 4.7112 & // language: Go \\
    \textbf{HTML} & 4.9355 & 3.1096 & <!--language: HTML--> \\
    \textbf{Shell} & 2.7498 & 1.7325 & \# language: Shell \\
    \textbf{PHP} & 2.1698 & 1.3671 & // language: PHP \\
    \textbf{CSS} & 1.5674 & 0.9876 & /* language: CSS */ \\
    \textbf{TypeScript} & 1.1667 & 0.7351 & // language: TypeScript \\
    \textbf{SQL} & 1.1533 & 0.7267 & -- language: SQL \\
    \textbf{TeX} & 0.8257 & 0.5202 & \% language: TeX\\
    \textbf{Rust} & 0.5228 & 0.3294 & // language: Rust \\
    \textbf{Objective-C} & 0.4526 & 0.2851 & // language: Objective-C \\
    \textbf{Scala} & 0.3786 & 0.2385 & // language: Scala \\
    \textbf{Kotlin} & 0.1707 & 0.1075 & // language: Kotlin \\
    \textbf{Pascal} & 0.0839 & 0.0529 & // language: Pascal \\
    \textbf{Fortran} & 0.077 & 0.0485 & !language: Fortran \\
    \textbf{R} & 0.0447 & 0.0281 & \# language: R \\
    \textbf{Cuda} & 0.0223 & 0.014 & // language: Cuda \\
    \textbf{C\#} & 0.0218 & 0.0138 & // language: C\# \\
    \textbf{Objective-C++} & 0.0014 & 0.0009 & // language: Objective-C++ \\
    \bottomrule
    \end{tabular}}
    }
\label{tab:dataset-comp}
\end{table}

\subsection{Tokenization of \name}
\label{app:tokenization}

\begin{figure}[htbp]
    \centering
    \includegraphics[width=.9\columnwidth]{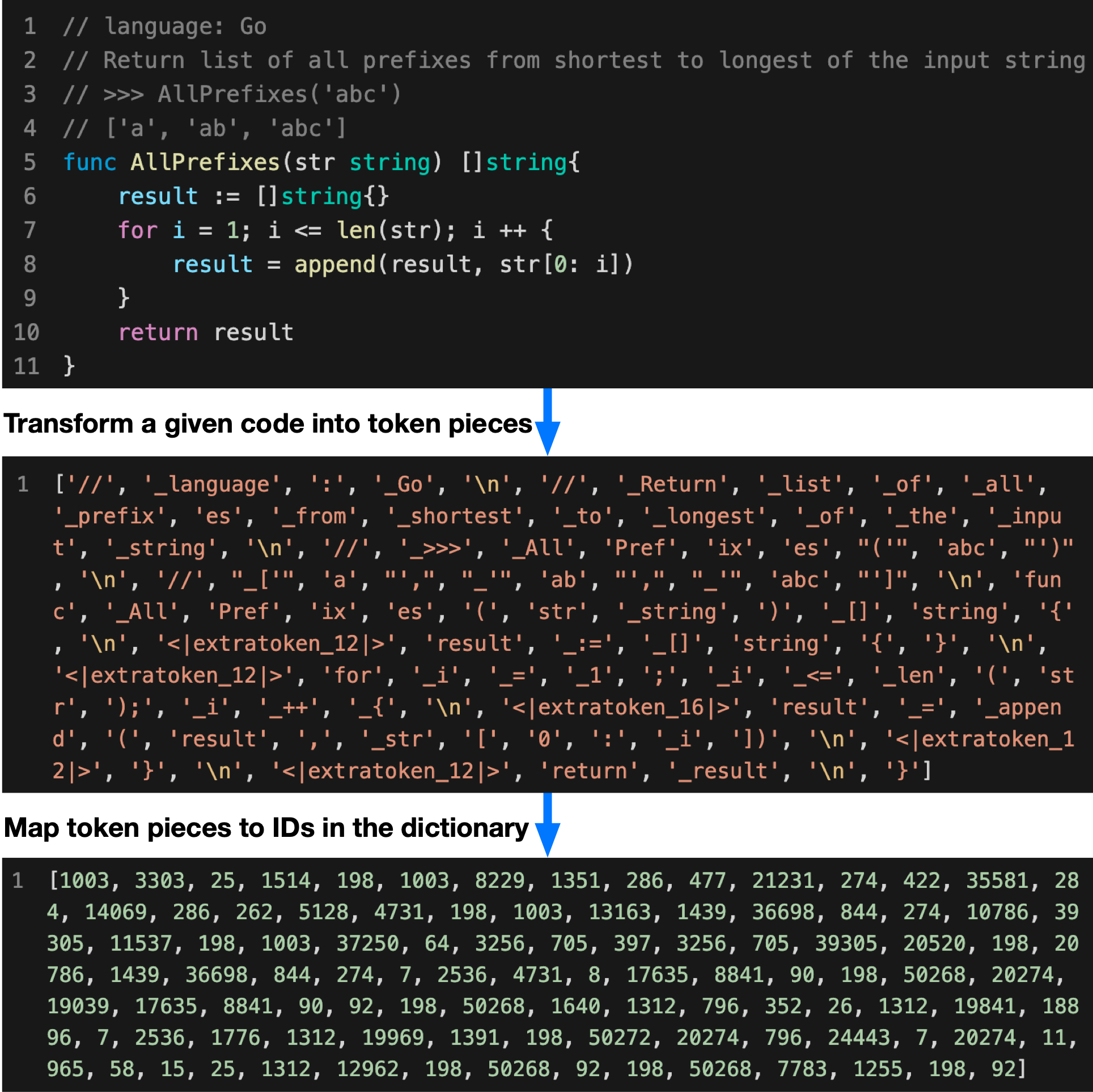}
    \caption{Illustration of tokenization in \name. \texttt{"\_"} represents a whitespace, and \texttt{"<|extratoken\_X|>"} represents concatenated whitespaces of different lengths.}
    \label{fig:tokenization}
\end{figure}

Given a code snippet as in~\figurename~\ref{fig:tokenization}, it is first separated into token pieces by the tokenizer.
Then, each token is mapped to an integer according to its ID in the pre-defined dictionary.
For example, 4 or 8 whitespaces (one or two indents in Python) are concatenated to \texttt{<|extratoken\_12|>} or \texttt{<|extratoken\_16|>}, respectively.
Note that in Figure~\figurename~\ref{fig:tokenization}, tokens are starting with \texttt{"\_"}, which represents whitespace and is often used to indicate if the token appears in the middle of a sentence. 
After tokenization, any code snippet or text description can be transformed into a vector of integers.

\subsubsection{Details of Budget Allocation Strategies}

We compare three strategies: \textbf{Best Single}, choose a single language with the best performance; \textbf{Uniform}, allocate the budget uniformly; \textbf{Weighted}, allocate the budget to multiple languages based their proportions in the training corpus. 
Detailed weights can be found in~\tablename~\ref{tab:app_alloc}.
The allocation of CodeGen-Multi-16B and InCoder-6.7B are extracted from the training corpora description in the original papers. 
The allocation of GPT-J-6B/GPT-NeoX-20B are from the number of tokens in the GitHub section of the Pile. 

\begin{table}[hbp]
\centering
\caption{Detailed assignment of budget allocation strategies. Given budget $k=100$, \textbf{Weighted} distribute the budgets according to the proportions of language in the training corpus of each model.}
\resizebox{.7\columnwidth}{!}{
\begin{tabular}{c|c|ccccc} 
\toprule
\textbf{Strategy}                           & \textbf{Model} & \textbf{Python} & \textbf{C++} & \textbf{Java} & \textbf{JavaScript} & \textbf{Go}  \\
\midrule
\textbf{Uniform}                   & All            & 20              & 20           & 20            & 20                  & 20           \\
\midrule
\multirow{5}{*}{\textbf{Weighted}} & GPT-J-6B          & 17              & 36           & 11            & 22                  & 14           \\
                                   & GPT-NeoX-20B       & 17              & 36           & 11            & 22                  & 14           \\
                                   & InCoder-6.7B        & 45              & 12           & 5             & 34                  & 4            \\
                                   & CodeGen-Multi-6B/16B        & 17              & 38           & 29            & 8                   & 8            \\
                                   & CodeGeeX-13B (ours)      & 32              & 33           & 20            & 9                   & 6            \\
\bottomrule
\end{tabular}}
\label{tab:app_alloc}
\end{table}

\subsection{Evaluation on \bench (Additional)}
\label{app:hx-additional}

\begin{figure}[htbp]
    \centering
    \begin{minipage}{0.42\textwidth}
    \includegraphics[width=\textwidth]{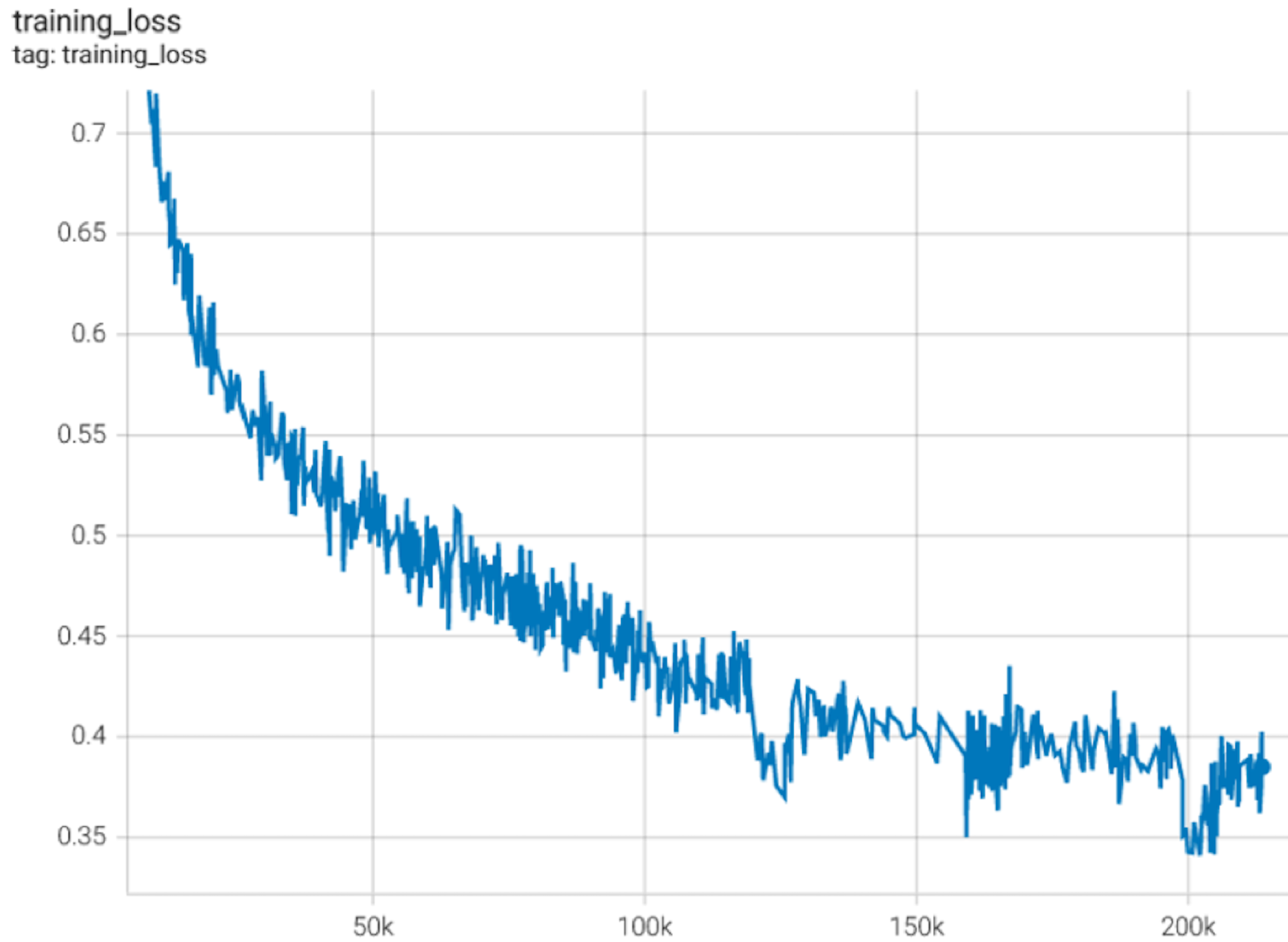}
    \captionof{figure}{Training loss of \name.}
    \label{fig:training-loss}
    \end{minipage}
    \begin{minipage}{0.52\textwidth}
    \includegraphics[width=0.95\textwidth]{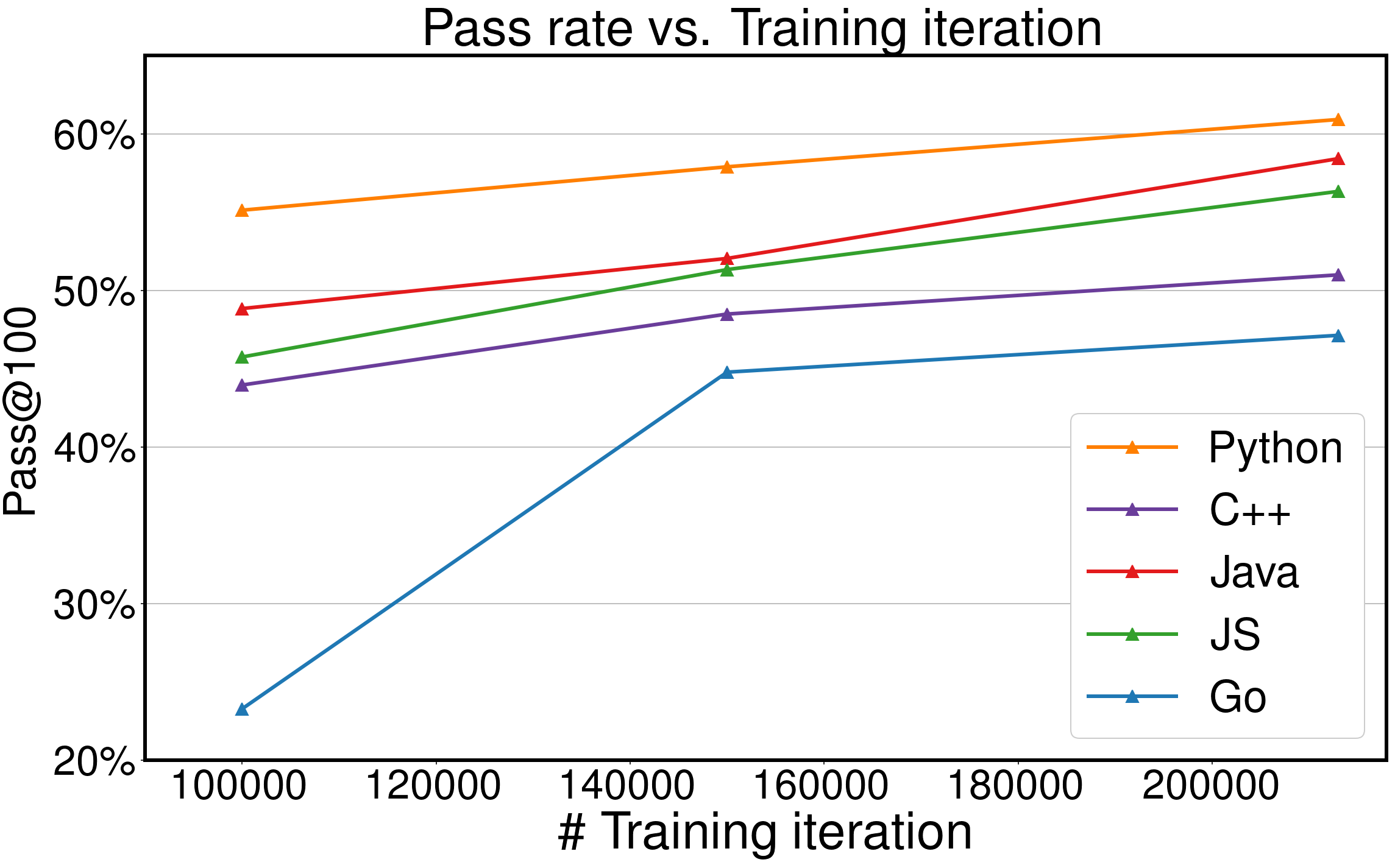}
    \captionof{figure}{\bench pass rate vs. iteration. } 
    \label{fig:hx-iter}
    \end{minipage}
\end{figure}

\textbf{Pass rate vs. number of training iterations.} We show in~\figurename~\ref{fig:training-loss} that the cross entropy loss decreases steadily during training while the pass rate in \bench continues to improve for different languages in~\figurename~\ref{fig:hx-iter}.

\begin{figure}[htbp]
    \centering
    \begin{minipage}[t]{0.5\textwidth}
    \centering
    \includegraphics[width=\textwidth]{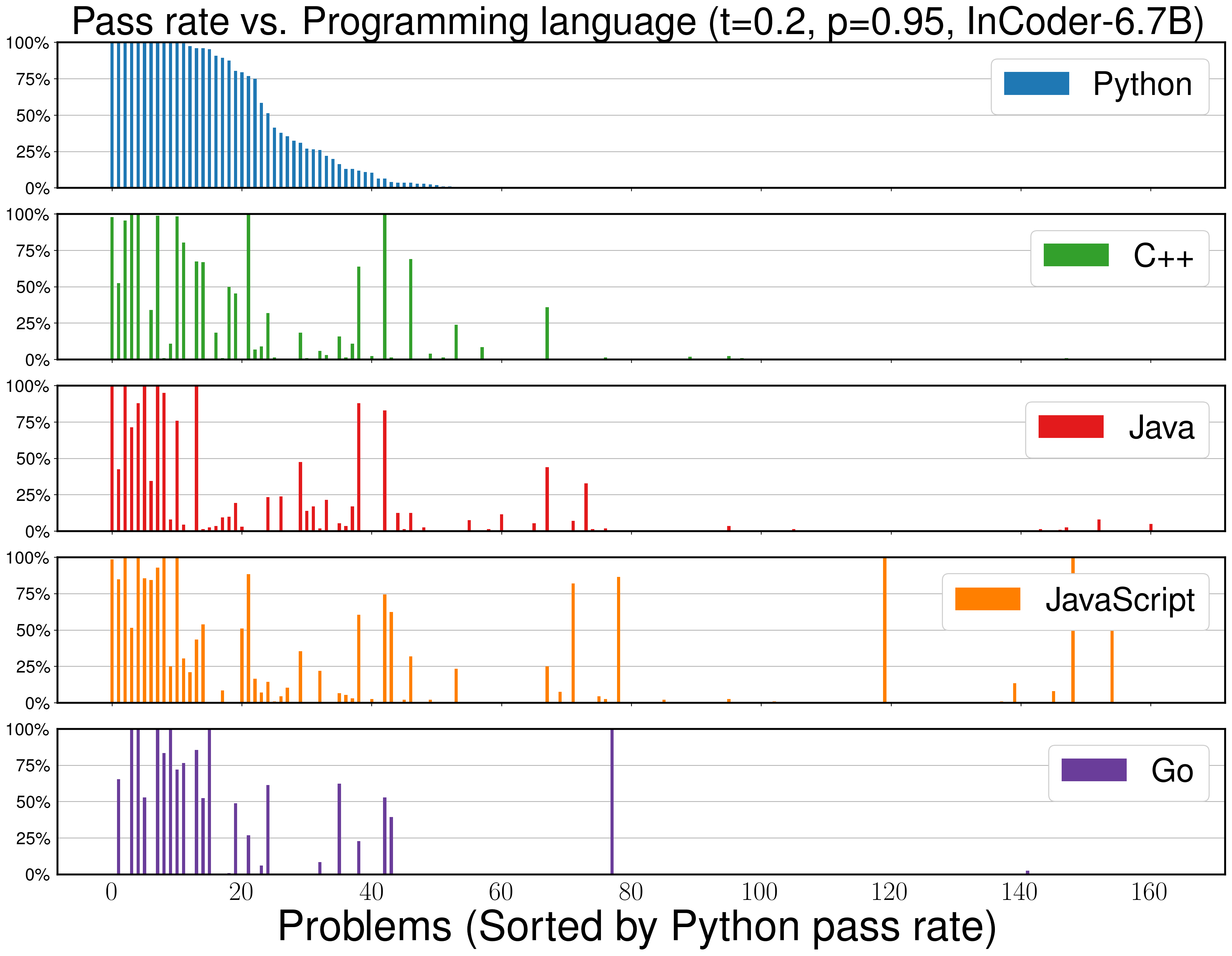}
    \end{minipage}%
    \begin{minipage}[t]{0.5\textwidth}
    \centering
    \includegraphics[width=\textwidth]{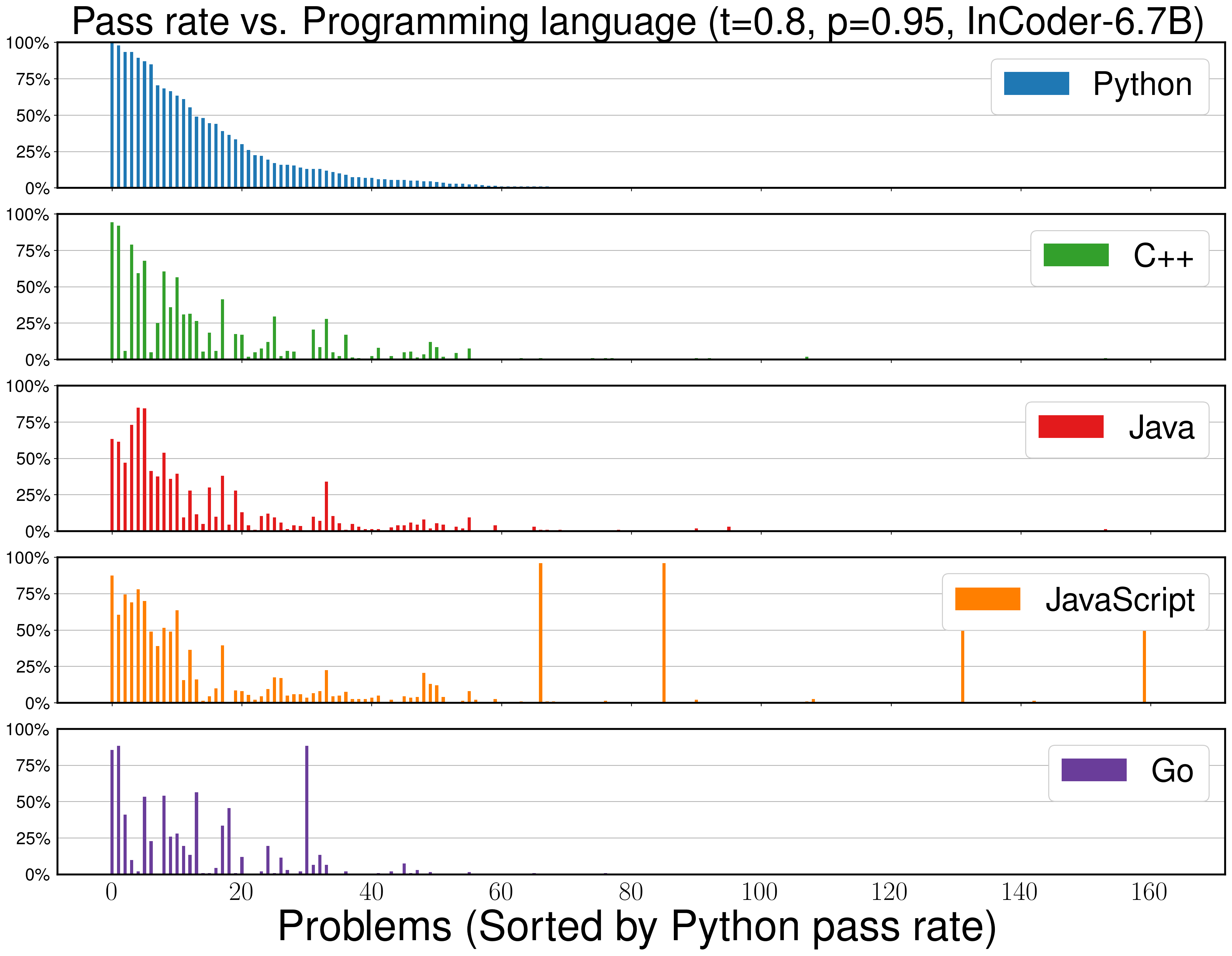}
    \end{minipage}
    \begin{minipage}[t]{0.5\textwidth}
    \centering
    \includegraphics[width=\textwidth]{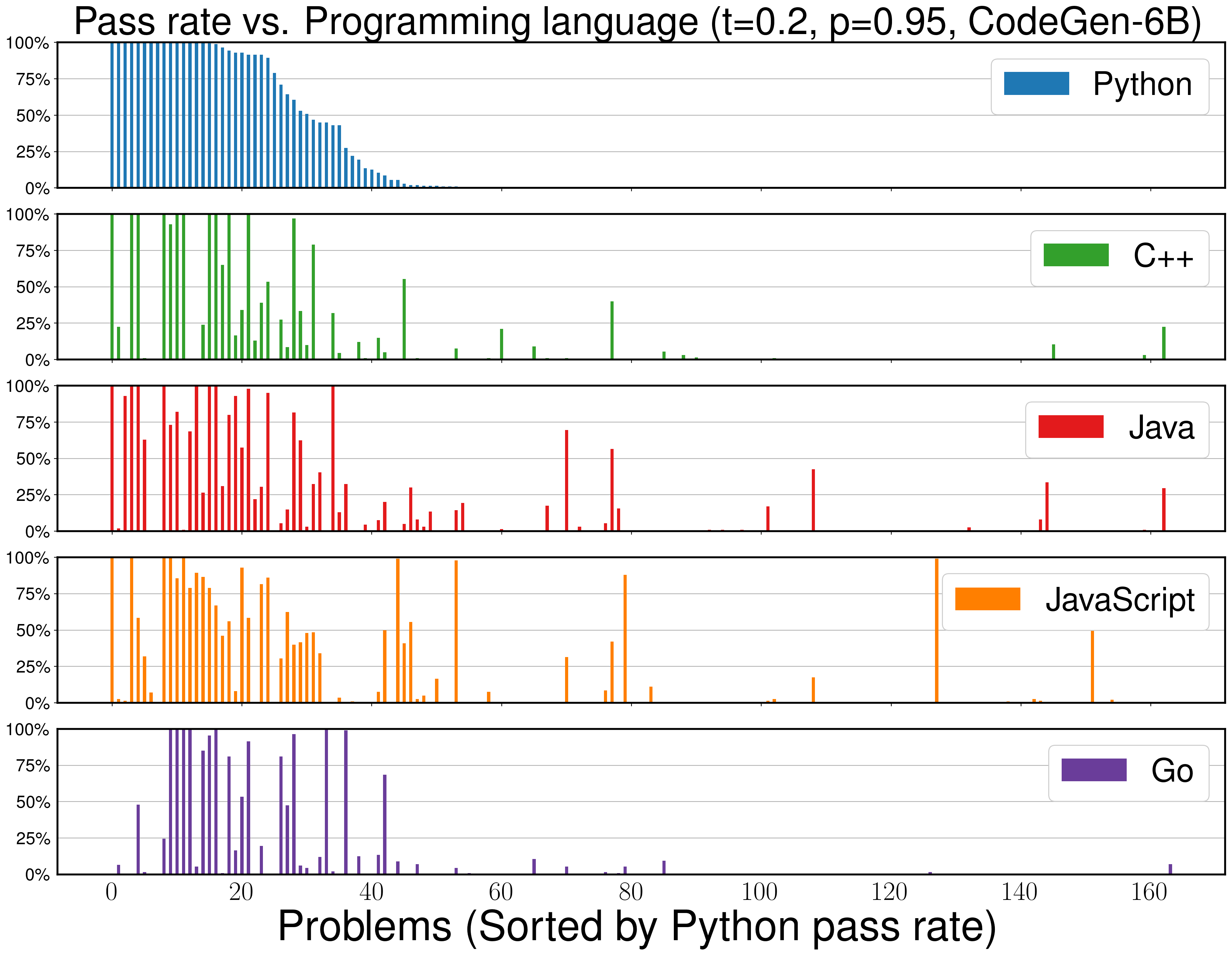}
    \end{minipage}%
    \begin{minipage}[t]{0.5\textwidth}
    \centering
    \includegraphics[width=\textwidth]{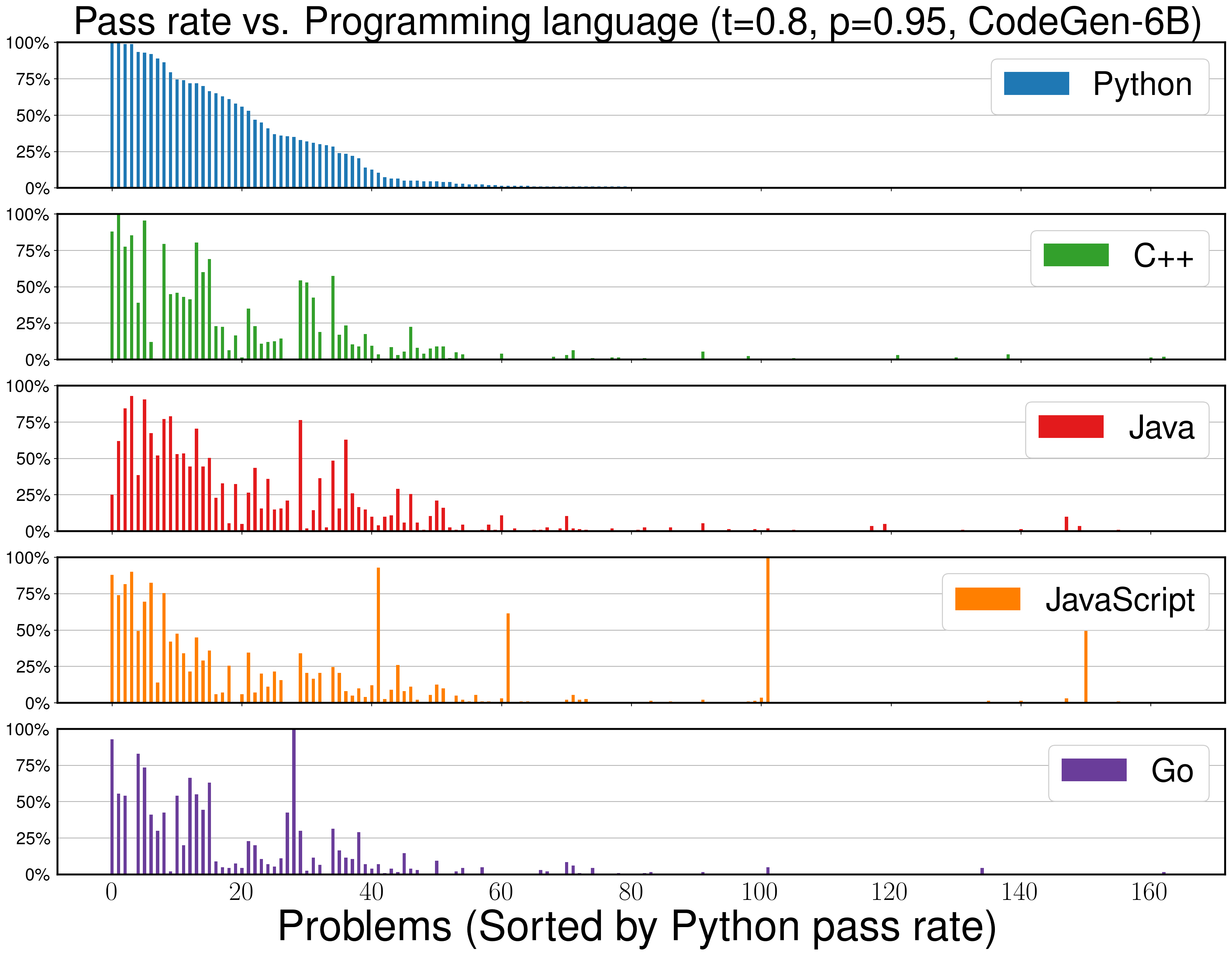}
    \end{minipage}
    \begin{minipage}[t]{0.5\textwidth}
    \centering
    \includegraphics[width=\textwidth]{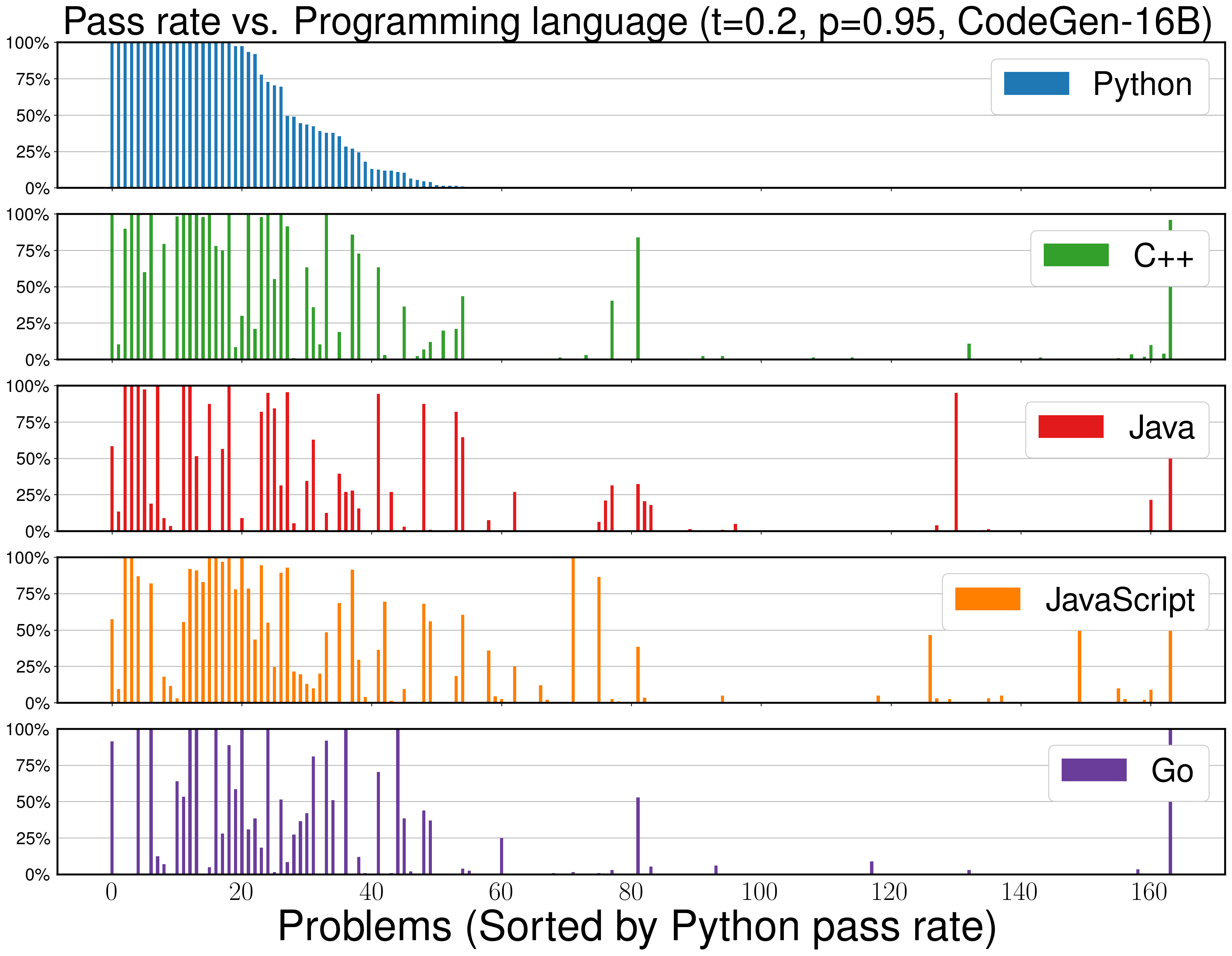}
    \end{minipage}%
    \begin{minipage}[t]{0.5\textwidth}
    \centering
    \includegraphics[width=\textwidth]{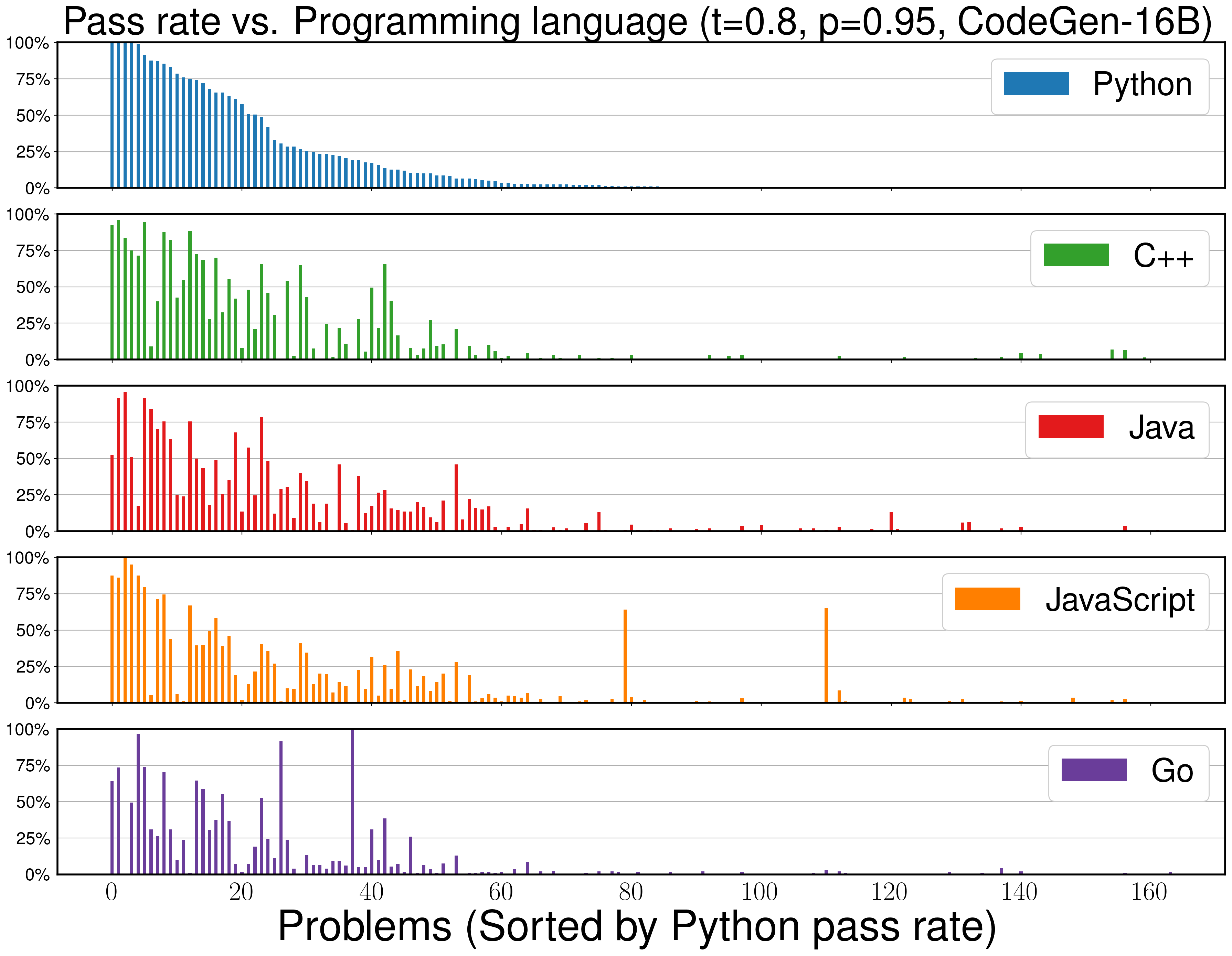}
    \end{minipage}
    \caption{In \bench, each problem's pass rate varies when generating in different programming languages. \textbf{Left:} $t=0.2, p=0.95$; \textbf{Right:} $t=0.8, p=0.95$. \textbf{From top to bottom}: InCoder-6.7B, CodeGen-Multi-6B, CodeGen-Multi-16B.}
    \label{fig:hx-distribution-other}
\end{figure}

\textbf{Pass rate distribution vs. Languages for other code generation models.} We show in~\figurename~\ref{fig:hx-distribution-other} that other code generation models also have various pass rate distribution for different languages.

\subsection{Evaluation on Other Benchmarks}
\subsubsection{Evaluation on HumanEval}

The evaluation setting on HumanEval is the same as \bench.
We show that among multilingual code generation models, \name achieves the second highest performance on HumanEval, reaching 60\% in pass@100 (surpassed by PaLMCoder-540B).
We also notice that monolingual models outperforms multilingual ones with a large margin, indicating that multilingual models require a larger model capacity to master different languages.

\begin{table}[hp]
    \centering
    \caption{The results of \name on HumanEval benchmark. The metric is pass@$k$ introduced in \cite{chen2021codex} (* use the biased pass@$k$ from \cite{chowdhery2022Palm}). Nucleus sampling is used with top-p=0.95 and sampling temperature being 0.2/0.6/0.8 for @1/@10/@100 respectively.}
    \resizebox{\columnwidth}{!}{
    \begin{tabular}{lcccccc}
    \toprule
    \textbf{Model} & \textbf{Size} & \textbf{Type} & \textbf{Available} & \textbf{pass@1} & \textbf{pass@10} & \textbf{pass@100} \\
    \midrule
    \textbf{CodeParrot}~\citep{tunstall2022codeparrot} & 1.5B & Multi & Yes & 4.00\% & 8.70\% & 17.90\% \\
    \textbf{PolyCoder}~\citep{xu2022polycoder} & 2.7B & Multi & Yes & 5.60\% & 9.80\% & 17.70\% \\
    \textbf{GPT-J}~\citep{wang2021gptj} & 6B & Multi & Yes & 11.60\% & 15.70\% & 27.70\% \\
    \textbf{CodeGen-Multi}~\citep{nijkamp2022codegen} & 6.1B & Multi & Yes & 18.16\% & 27.81\% & 44.85\% \\
    \textbf{InCoder}~\citep{fried2022incoder} & 6.7B & Multi & Yes & 15.20\% & 27.80\% & 47.00\% \\
    \textbf{GPT-NeoX}~\citep{black2022gptneox} & 20B & Multi & Yes & 15.40\% & 25.60\% & 41.20\% \\
    \textbf{LaMDA}~\citep{thoppilan2022lamda} & 137B & Multi & No & 14.00\%* & - & 47.30\%* \\
    \textbf{CodeGen-Multi}~\citep{nijkamp2022codegen} & 16.1B & Multi & Yes & 19.22\% & 34.64\% & 55.17\% \\
    \textbf{PaLM-Coder}~\citep{chowdhery2022Palm} & 540B & Multi & No & \textbf{36.00\%}* & - & \textbf{88.40\%}* \\
    \midrule
    \textbf{Codex}~\citep{chen2021codex} & 12B & Mono & No & 28.81\% & 46.81\% & 72.31\% \\
    \textbf{CodeGen-Mono}~\citep{nijkamp2022codegen} & 16.1B & Mono & Yes & 29.28\% & \textbf{49.86\%} & 75.00\% \\
    \midrule
    \textbf{\name (ours)} & 13B & Multi & Yes & 22.89\% & 39.57\% & 60.92\% \\
    \bottomrule
    \end{tabular}}
    \label{tab:humaneval}
\end{table}

\subsubsection{Evaluation on MBPP}

MBPP dataset is proposed by~\cite{austin2021program}, containing 974 problems in Python. 
Due to specific input-output format, MBPP need to be evaluated under a few-shot setting.
We follow the splitting in the original paper and use problems 11-510 for testing.
Under 1-shot setting, we use problem 2 in prompts.
Under 3-shot setting, we use problem 2,3,4 in prompts.
The metric is pass@$k$, $k\in\{1, 10, 80\}$.
For pass@1, the temperature is 0.2 and top-p is 0.95; for pass@10 and pass@ 80, the temperature is 0.8 and top-p is 0.95.
For baselines, we consider LaMDA-137B, PaLM-540B, Code-davinci-002 (online API version of OpenAI Codex), PaLMCoder-540B and InCoder-6.7B.

The results indicate that the model capacity is essential for multilingual code generation model.
With significantly more parameters, PaLM and Codex outperform \name with a large margin.
Meanwhile, we find that more shot in the prompts harm the performance of \name, the same phenomenon have also been discovered in InCoder~\citep{fried2022incoder}.
We assume that it is because smaller models do not have enough reasoning ability to benefit from the few-shot setting.

\begin{table}
\centering
\caption{The results of \name on MBPP dataset~\citep{austin2021program}. }
\resizebox{.9\columnwidth}{!}{
\begin{tabular}{clcccc}
\toprule
\textbf{Method} & \multicolumn{1}{c}{\textbf{Model}} & \textbf{Pass@1} & \textbf{Pass@10} & \textbf{Pass@80} \\
\midrule
\multirow{5}{*}{3-shot} & \textbf{LaMDA-137B}~\citep{austin2021program} & 14.80 & - & 62.40 \\
 & \textbf{PaLM-540B}~\citep{chowdhery2022Palm} & 36.80 & - & 75.00 \\
 & \textbf{Code-davinci-002}~\citep{chen2021codex} & \textbf{50.40} & - & \textbf{84.40} \\
 & \textbf{PaLMCoder-540B}~\citep{chowdhery2022Palm} & 47.00 & - & 80.80 \\
 & \textbf{CodeGeeX-13B (ours)} & 22.44 & 43.24 & 63.52 \\
\midrule
\multirow{2}{*}{1-shot} & \textbf{InCoder-6.7B}~\citep{fried2022incoder} & 19.40 & - & - \\
 & \textbf{CodeGeeX-13B (ours)} & 24.37 & 47.95 & 68.50 \\
\bottomrule
\end{tabular}
}
\label{tab:app_mbpp}
\end{table}

\subsubsection{Evaluation on CodeXGLUE}
CodeXGLUE is a benchmark proposed by \cite{lu2021codexglue}, containing multiple datasets to support evaluation on multiple tasks, using similarity-based metrics like CodeBLEU, BLEU and accuracy for generation tasks.
We test the performance of CodeGeeX on the \textbf{code summarization} task of CodeXGLUE. 
We first fine-tune the parameters of CodeGeeX on the given training set, mixing the training data in all languages to get one fine-tuned model. Then, we test the performance of the fine-tuned model on each language, using BLEU score for evaluation because the models generate natural language in summarization tasks.

For all languages, we set temperature to 0.2 and top-p to 0.95, and generate one summarization for each sample in the test set. 
We report the results in \tableautorefname~\ref{tab:cxg-summary}. 
CodeGeeX obtains an average BLEU score of 20.63, besting all baseline models. It is worth noting that CodeGeeX is not pre-trained on Ruby, and after removing the results on Ruby for all models, CodeGeeX outperforms the best baseline model (DistillCodeT5 from \cite{wang2021codet5}) by 1.88 in the average BLEU score.

\begin{table}[ht]
    \centering
    \renewcommand{\arraystretch}{1.3}
    \caption{The results of \name on \textbf{code summarization} in CodeXGLUE benchmark~\citep{lu2021codexglue}. Six languages are considered, Ruby, JavaScript, Go, Python, Java, PHP. The metric is the BLEU score. * We don't have Ruby in the pretraining corpus.}
    \label{tab:cxg-summary}
    \resizebox{\columnwidth}{!}{
    \begin{tabular}{lccccccc} 
    \toprule
    \textbf{Model} & \textbf{All} & \textbf{Ruby} & \textbf{JavaScript} & \textbf{Go} & \textbf{Python} & \textbf{Java} & \textbf{PHP} \\
    \midrule
    \textbf{CodeBERT}~\citep{feng2020codebert} & 17.83 & 12.16 & 14.90 & 18.07 & 19.06 & 17.65 & 25.16 \\
    \textbf{PLBART}~\citep{ahmad2021plbart} & 18.32 & 14.11 & 15.56 & 18.91 & 19.30 & 18.45 & 23.58 \\
    \textbf{ProphetNet-X}~\citep{qi2021prophetnet} & 18.54 & 14.37 & \textbf{16.60} & 18.43 & 17.87 & 19.39 & 24.57 \\
    \textbf{CoTexT}~\citep{phan2021cotext} & 18.55 & 14.02 & 14.96 & 18.86 & 19.73 & 19.06 & 24.68 \\
    \textbf{PolyglotCodeBERT}~\citep{feng2020codebert} & 19.06 & 14.75 & 15.80 & 18.77 & 18.71 & 20.11 & 26.23 \\
    \textbf{DistillCodeT5}~\citep{wang2021codet5} & 20.01 & \textbf{15.75} & 16.42 & 20.21 & 20.59 & \textbf{20.51} & 26.58 \\
    \textbf{\name (ours)} & \textbf{20.63} & 10.05* & 16.01 & \textbf{24.62} & \textbf{22.50} & 19.60 & \textbf{31.00} \\
    \bottomrule
    \end{tabular}
    }
\end{table}

\subsubsection{Evaluation on XLCoST}
XLCoST is a benchmark proposed by \cite{zhu2022xlcost}, containing parallel multilingual code data, with code snippets aligned among different languages.
For generation tasks, XLCoST uses CodeBLEU, BLEU for evaluation.
We choose the \textbf{code translation} task of XLCoST for \name evaluation. 
We first fine-tune the parameters of CodeGeeX on the given training set, combining the training data in all 42 languages pairs to obtain one fine-tuned model. Then, we test the performance of the fine-tuned model on each language pair with CodeBLEU score.

For all language pairs, we set temperature to 0.2 and top-p to 0.95, and generate one translation for each sample in the test set. 
We report the results in \tableautorefname~\ref{tab:xlc-trans}. 
CodeGeeX performs better than all baseline models on all language pairs except for: PHP to Python on program level, C++ to Python on snippet level, and PHP to Python on snippet level. On average, CodeGeeX outperforms the baseline by 4.10 on program level and by 1.99 on snippet level.

\begin{table}[ht]
    \centering
    \renewcommand{\arraystretch}{1.3}
    \caption{The results of \name on \textbf{code translation} in XLCoST benchmark. Six languages are considered, C++, Java, Python, C\#, JavaScript, PHP, C. The metric is CodeBLEU \citep{ren2020codebleu}. The results of baselines are adopted from the original paper~\citep{zhu2022xlcost}.}
    \label{tab:xlc-trans}
    \resizebox{\columnwidth}{!}{
    \begin{tabular}{ccccccccc|ccccccc} 
    \toprule
     &  & \multicolumn{7}{c}{\textbf{Snippet-level}} & \multicolumn{7}{c}{\textbf{Program-level}} \\
     & \textbf{Model} & \textbf{C++} & \textbf{Java} & \textbf{Py} & \textbf{C\#} & \textbf{JS} & \textbf{PHP} & \textbf{C} & \textbf{C++} & \textbf{Java} & \textbf{Py} & \textbf{C\#} & \textbf{JS} & \textbf{PHP} & \textbf{C} \\
     \midrule
    \multirow{4}{*}{\textbf{C++}} & CodeBERT & – & 84.94 & 74.55 & 84.99 & 82.79 & 68.56 & 45.46 & – & 74.73 & 24.96 & 76.35 & 72.95 & 50.40 & 21.84 \\
     & PLBART & – & 83.85 & 74.89 & 84.57 & 83.19 & 68.62 & 83.95 & – & 75.26 & 70.13 & 78.01 & 61.85 & 67.01 & 72.59 \\
     & CodeT5 & – & 86.35 & \textbf{76.28} & 85.85 & 84.31 & 69.87 & 90.45 & – & 80.03 & 71.56 & 81.73 & 79.48 & 70.44 & 85.67 \\
     & \textbf{\name} & – & \textbf{86.99} & 74.73 & \textbf{86.63} & \textbf{84.83} & \textbf{70.30} & \textbf{94.04} & – & \textbf{84.40} & \textbf{73.89} & \textbf{84.49} & \textbf{82.20} & \textbf{71.18} & \textbf{87.32} \\
     \midrule
    \multirow{4}{*}{\textbf{Java}} & CodeBERT & 87.27 & – & 58.39 & 92.26 & 84.63 & 67.26 & 39.94 & 79.36 & – & 8.51 & 84.43 & 76.02 & 51.42 & 21.22 \\
     & PLBART & 87.31 & – & 58.30 & 90.78 & 85.42 & 67.44 & 72.47 & 81.41 & – & 66.29 & 83.34 & 80.14 & 67.12 & 63.37 \\
     & CodeT5 & 88.26 & – & 74.59 & 92.56 & 86.22 & 69.02 & 82.78 & 84.26 & – & 69.57 & 87.79 & 80.67 & 69.44 & 78.78 \\
     & \textbf{\name} & \textbf{89.08} & – & \textbf{74.65} & \textbf{92.94} & \textbf{86.96} & \textbf{69.77} & \textbf{88.44} & \textbf{87.07} & – & \textbf{73.11} & \textbf{91.78} & \textbf{84.34} & \textbf{70.61} & \textbf{81.07} \\
     \midrule
    \multirow{4}{*}{\textbf{Py}} & CodeBERT & 80.46 & 58.50 & – & 54.72 & 57.38 & 65.14 & 10.70 & 68.87 & 28.22 & – & 17.80 & 23.65 & 49.30 & 18.32 \\
     & PLBART & 80.15 & 74.15 & – & 73.50 & 73.20 & 66.12 & 62.15 & 74.38 & 67.80 & – & 66.03 & 69.30 & 64.85 & 29.05 \\
     & CodeT5 & 81.56 & 78.61 & – & 78.89 & 77.76 & 67.54 & 68.67 & 78.85 & 73.15 & – & 73.35 & 71.80 & 67.50 & 56.35 \\
     & \textbf{\name} & \textbf{82.91} & \textbf{81.93} & – & \textbf{81.30} & \textbf{79.83} & \textbf{67.99} & \textbf{82.59} & \textbf{82.49} & \textbf{79.03} & – & \textbf{80.01} & \textbf{77.47} & \textbf{68.91} & \textbf{71.67} \\
     \midrule
    \multirow{4}{*}{\textbf{C\#}} & CodeBERT & 86.96 & 90.15 & 56.92 & – & 84.38 & 67.18 & 40.43 & 78.52 & 82.25 & 10.82 & – & 75.46 & 51.76 & 21.63 \\
     & PLBART & 84.98 & 6.27 & 69.82 & – & 85.02 & 67.30 & 75.74 & 80.17 & 81.37 & 67.02 & – & 79.81 & 67.12 & 57.60 \\
     & CodeT5 & 88.06 & 91.69 & 73.85 & – & 85.95 & 68.97 & 81.09 & 83.59 & 85.70 & 69.52 & – & 80.50 & 69.63 & 77.35 \\
     & \textbf{\name} & \textbf{88.70} & \textbf{93.03} & \textbf{74.55} & – & \textbf{86.44} & \textbf{69.49} & \textbf{86.69} & \textbf{87.11} & \textbf{90.46} & \textbf{72.89} & – & \textbf{83.83} & \textbf{70.58} & \textbf{80.73} \\
     \midrule
    \multirow{4}{*}{\textbf{JS}} & CodeBERT & 84.38 & 84.42 & 52.57 & 84.74 & – & 66.66 & 33.29 & 75.43 & 72.33 & 9.19 & 75.47 & – & 52.08 & 19.79 \\
     & PLBART & 84.45 & 84.90 & 69.29 & 85.05 & – & 67.09 & 72.65 & 80.19 & 76.96 & 64.18 & 78.51 & – & 67.24 & 67.70 \\
     & CodeT5 & 85.06 & 85.48 & 73.15 & 85.96 & – & 68.42 & 80.49 & 82.14 & 79.91 & 68.42 & 81.77 & – & 68.76 & 74.57 \\
     & \textbf{\name} & \textbf{86.72} & \textbf{86.96} & \textbf{73.25} & \textbf{86.41} & – & \textbf{69.00} & \textbf{83.85} & \textbf{85.84} & \textbf{83.85} & \textbf{72.11} & \textbf{85.35} & – & \textbf{69.80} & \textbf{79.41} \\
     \midrule
    \multirow{4}{*}{\textbf{PHP}} & CodeBERT & 82.58 & 81.57 & 69.29 & 80.96 & 79.94 & – & 28.45 & 50.13 & 46.81 & 16.92 & 49.75 & 48.12 & – & 22.19 \\
     & PLBART & 83.87 & 81.66 & 71.17 & 78.00 & 82.94 & – & 57.39 & 79.40 & 72.77 & 61.26 & 74.16 & 44.26 & – & 56.23 \\
     & CodeT5 & 86.33 & 85.12 & \textbf{73.22} & 84.56 & 83.56 & – & 79.30 & 85.55 & 82.09 & \textbf{72.26} & 83.79 & 81.72 & – & 65.86 \\
     & \textbf{\name} & \textbf{86.75} & \textbf{86.24} & 71.37 & \textbf{85.58} & \textbf{84.17} & – & \textbf{83.89} & \textbf{87.23} & \textbf{83.90} & 71.02 & \textbf{85.34} & \textbf{82.81} & – & \textbf{78.76} \\
     \midrule
    \multirow{4}{*}{\textbf{C}} & CodeBERT & 45.84 & 39.69 & 13.55 & 39.71 & 29.85 & 38.88 & – & 21.70 & 21.27 & 21.10 & 19.50 & 15.64 & 31.71 & – \\
     & PLBART & 82.53 & 72.35 & 49.16 & 75.78 & 75.05 & 60.86 & – & 78.42 & 13.45 & 5.53 & 45.15 & 31.47 & 25.17 & – \\
     & CodeT5 & 90.26 & 81.81 & 63.81 & 83.05 & 79.73 & 66.32 & – & 88.17 & 76.12 & 56.32 & 80.20 & 76.50 & 64.28 & – \\
     & \textbf{\name} & \textbf{91.30} & \textbf{85.58} & \textbf{71.52} & \textbf{87.52} & \textbf{84.91} & \textbf{68.52} & – & \textbf{88.21} & \textbf{82.46} & \textbf{69.78} & \textbf{85.56} & \textbf{81.21} & \textbf{68.80} & – \\
    \bottomrule
    \end{tabular}}
\end{table}

\clearpage

\subsection{Examples of \name Generation}
\label{app:example}

\begin{figure}[htbp]
    \centering
    \includegraphics[width=0.9\textwidth]{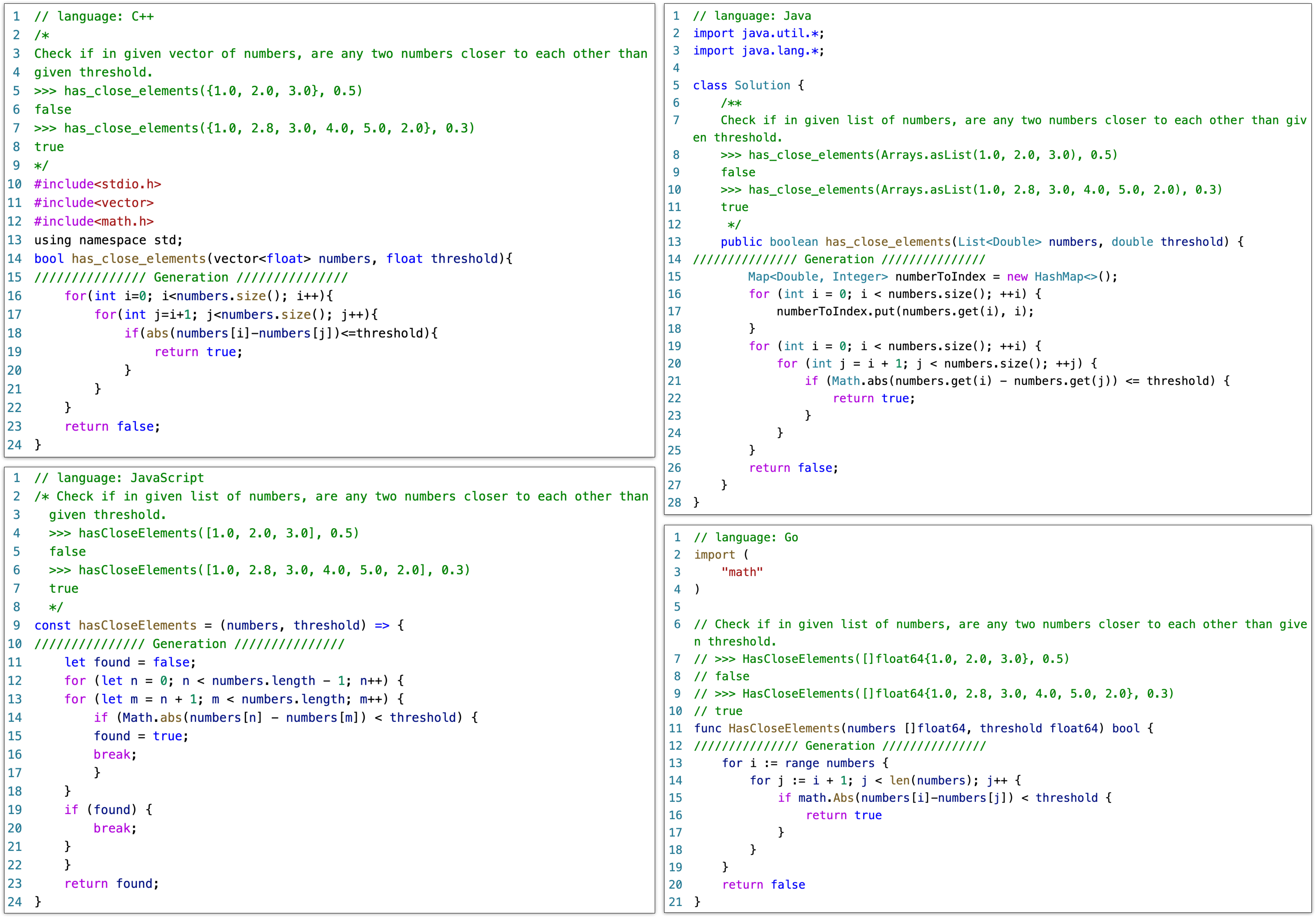}
    \caption{Solutions (Problem 0 in \bench) generated by \name. Prompt and generated codes are separated by the 'Generation' line (added after the generation as an indicator).}
    \label{fig:hx_example}
    \vspace{-5mm}
\end{figure}

\begin{figure}[htbp]
    \centering
    \includegraphics[width=0.9\textwidth]{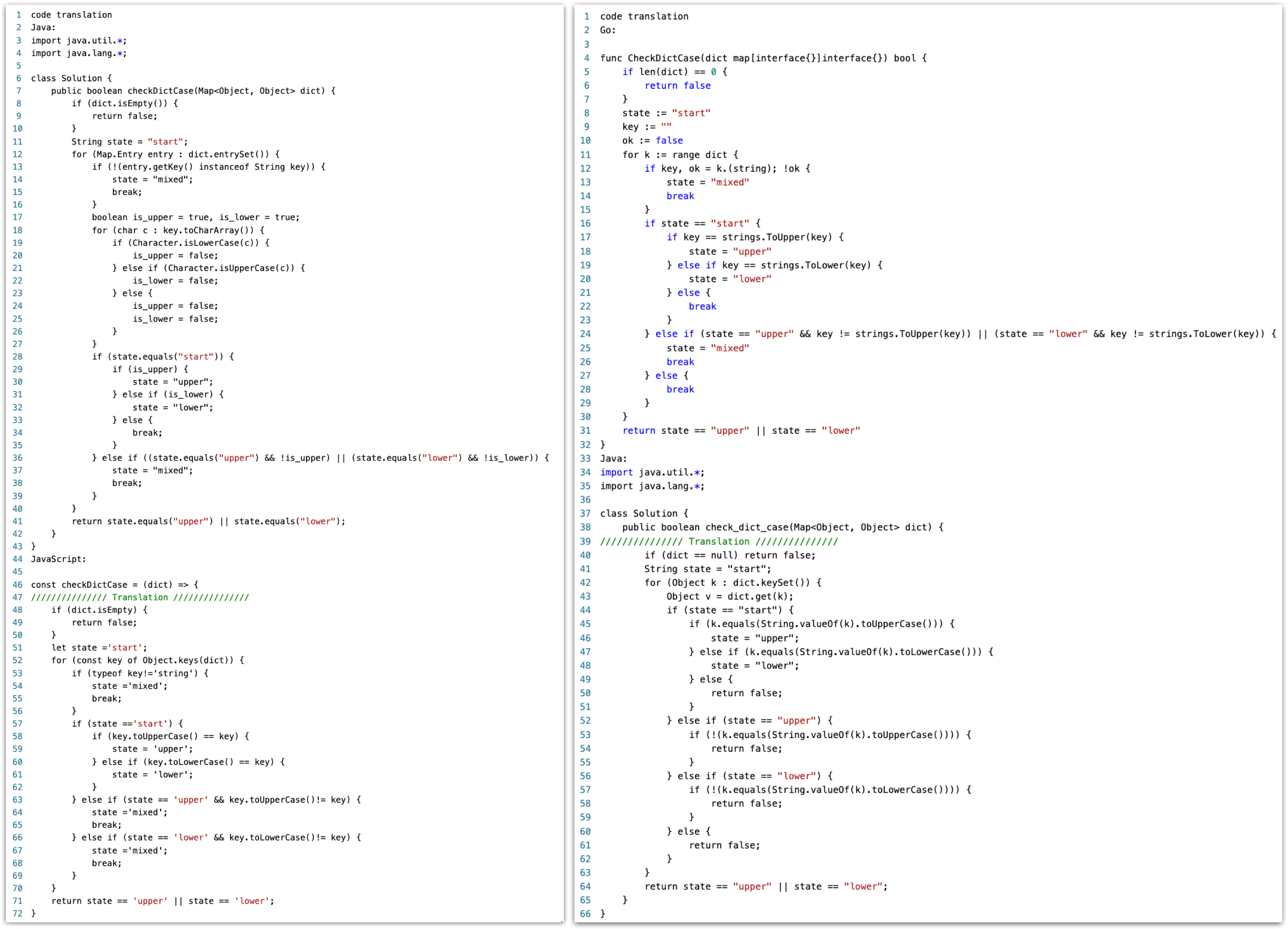}
    \caption{Solutions (Problem 95 in \bench) translated by \name. Prompt and generated codes are separated by the 'Translation' line (added after the generation as an indicator).}
    \label{fig:hx_example_trans_id_95}
    \vspace{-5mm}
\end{figure}

\begin{figure}[htbp]
    \centering
    \begin{minipage}[t]{\textwidth}
    \includegraphics[width=\textwidth]{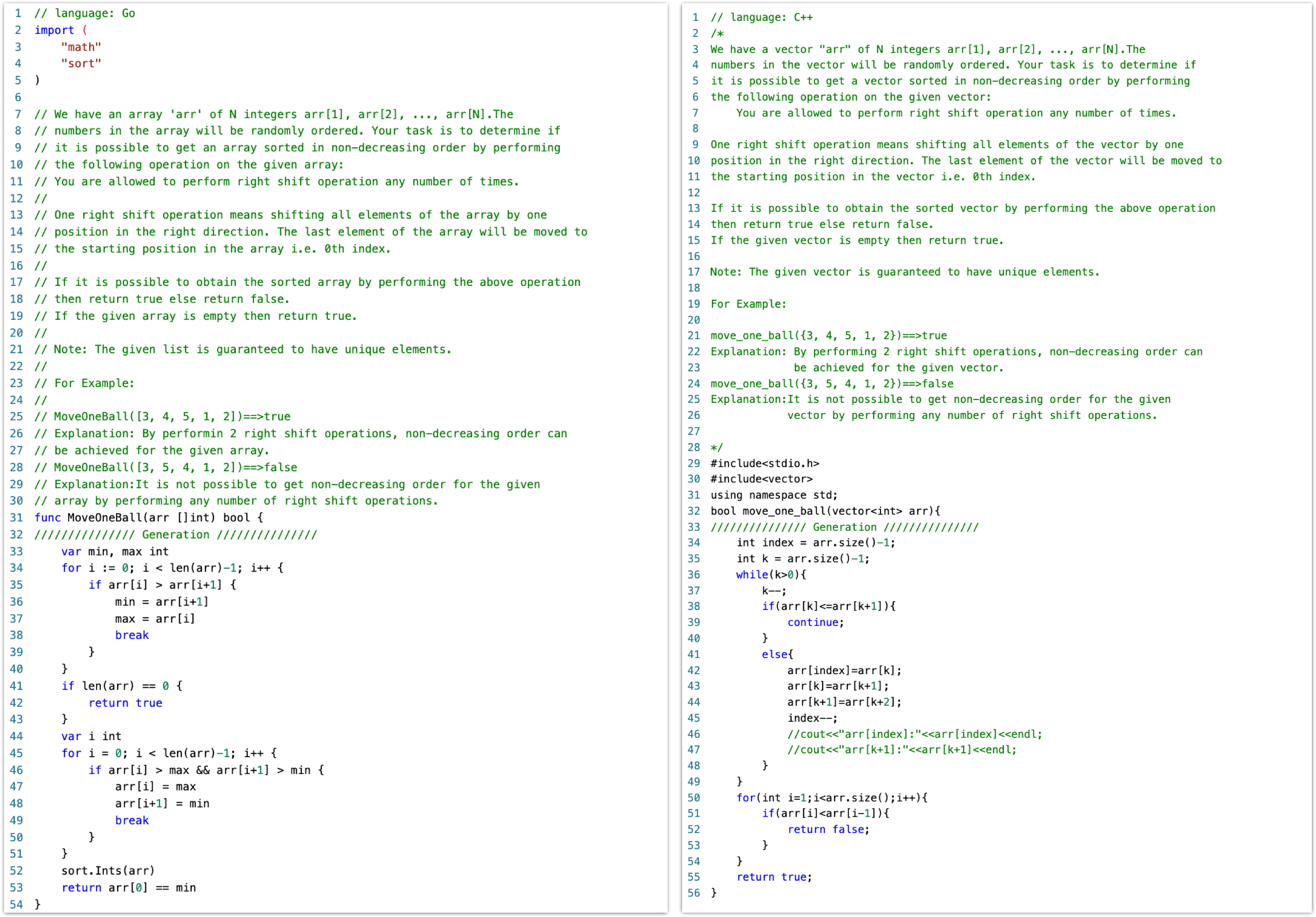}
    \end{minipage}
    \begin{minipage}[t]{\textwidth}
    \includegraphics[width=\textwidth]{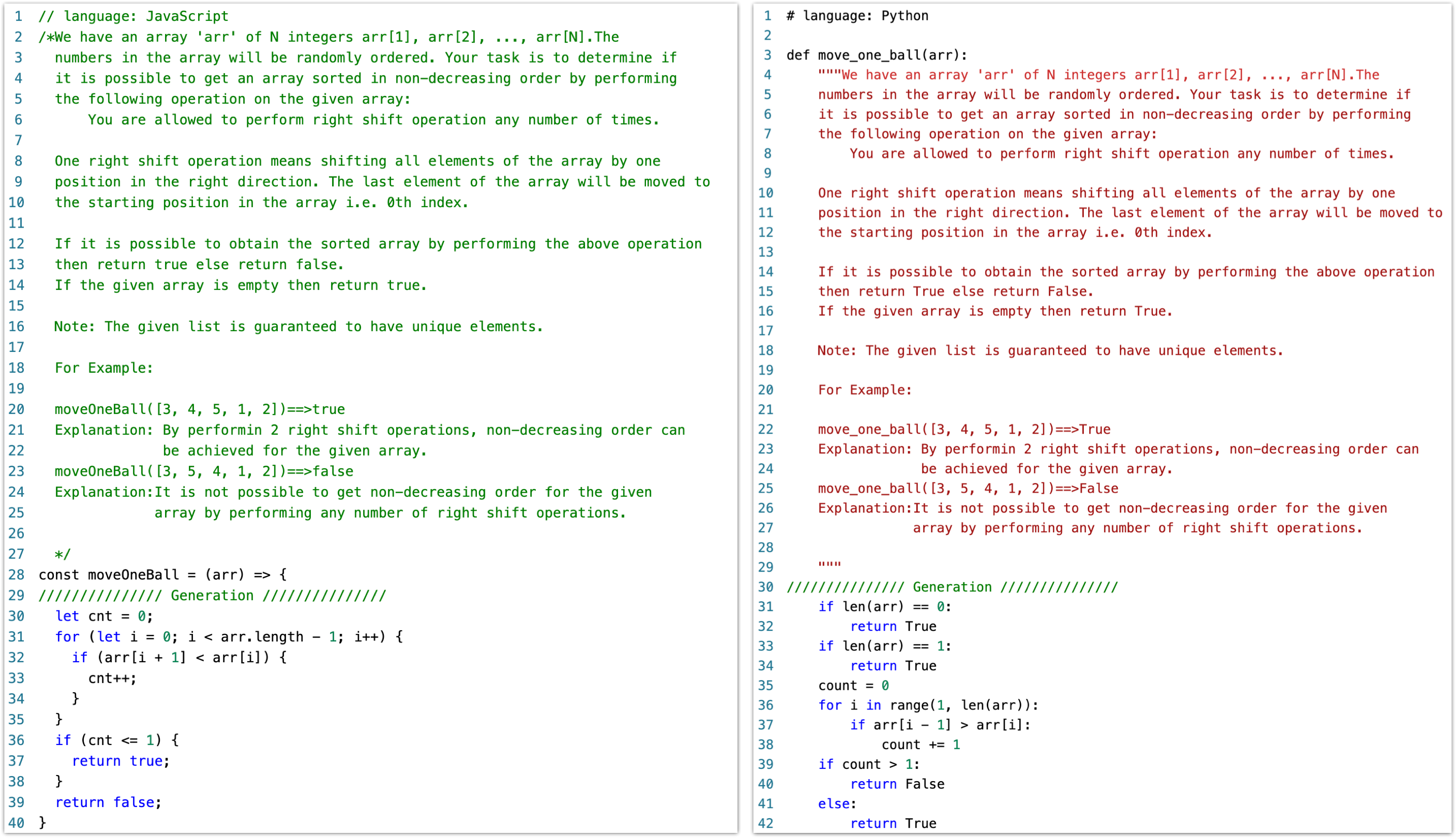}
    \end{minipage}
    \caption{Solutions (Problem 109 in \bench) generated by \name. Prompt and generated codes are separated by the 'Generation' line (added after the generation as an indicator).}
    \label{fig:hx_example_id_109}
\end{figure}

\begin{figure}[htbp]
    \centering
    \includegraphics[width=\textwidth]{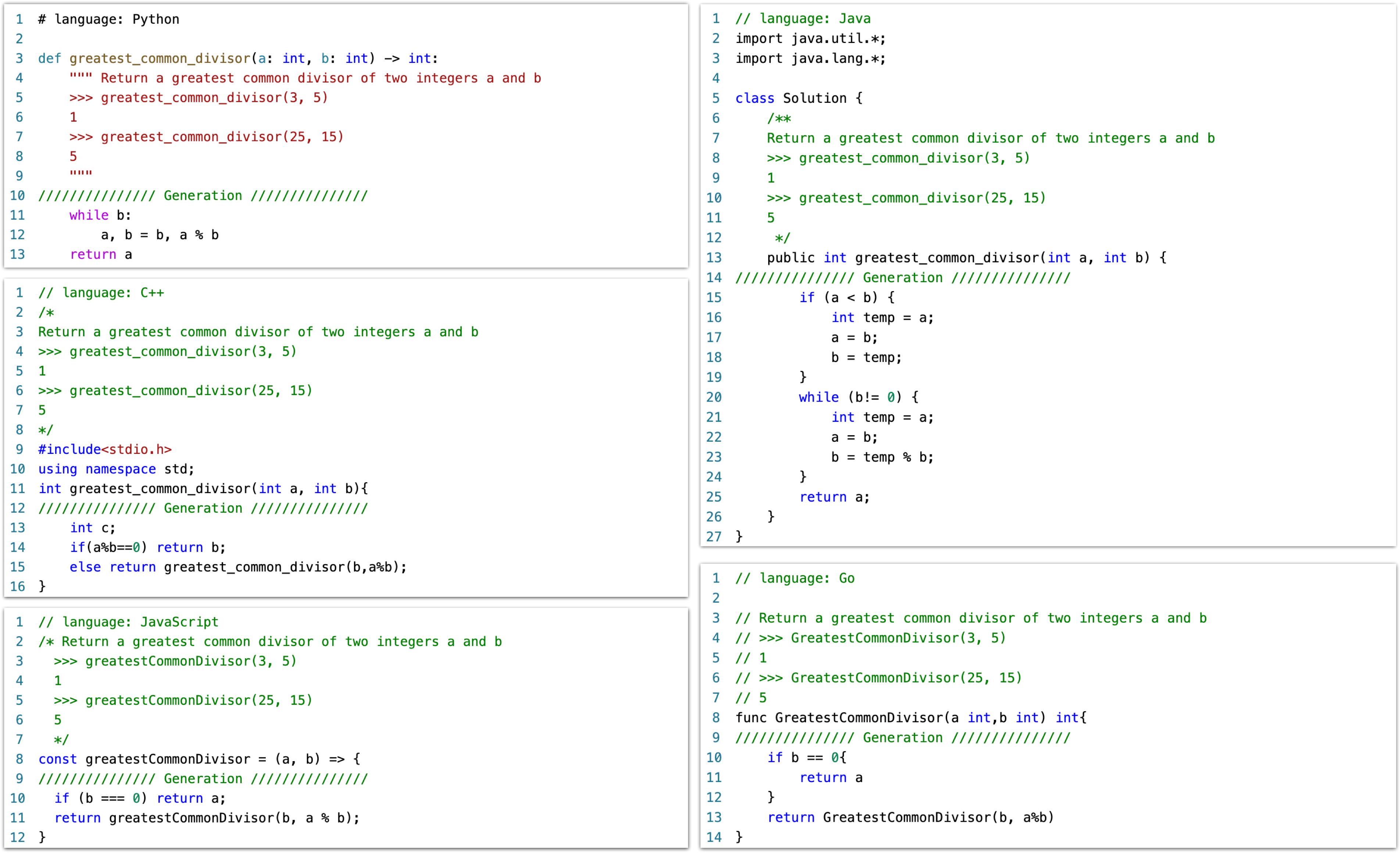}
    \caption{Solutions (Problem 13 in \bench) generated by \name. Prompt and generated codes are separated by the 'Generation' line (added after the generation as an indicator).}
    \label{fig:hx_example_id_13}
    \vspace{-5mm}
\end{figure}

\begin{figure}[htbp]
    \centering
    \includegraphics[width=\textwidth]{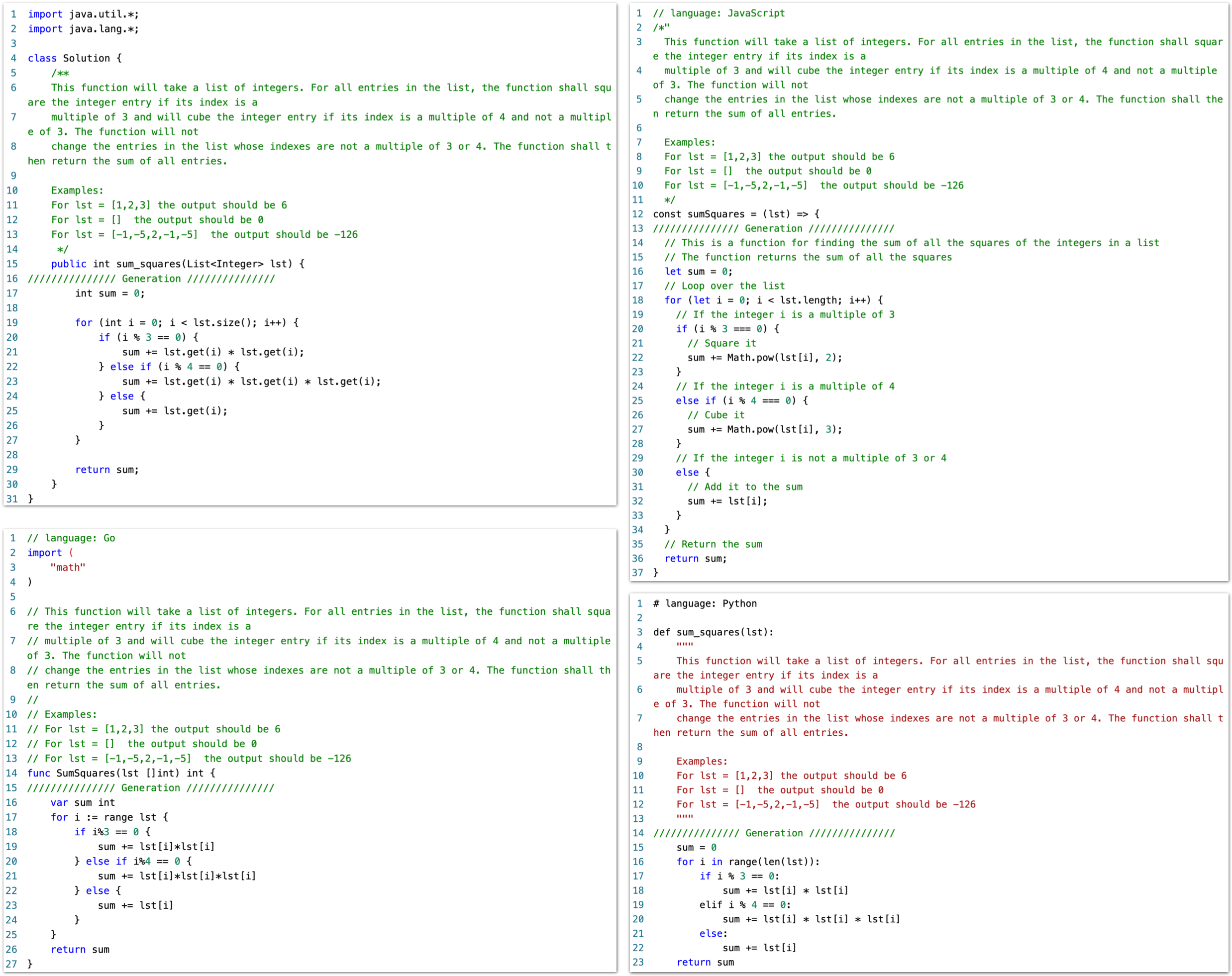}
    \caption{Solutions (Problem 142 in \bench) generated by \name. Prompt and generated codes are separated by the 'Generation' line (added after the generation as an indicator).}
    \label{fig:hx_example_id_142}
    \vspace{-5mm}
\end{figure}

\begin{figure}[htbp]
    \centering
    \includegraphics[width=\textwidth]{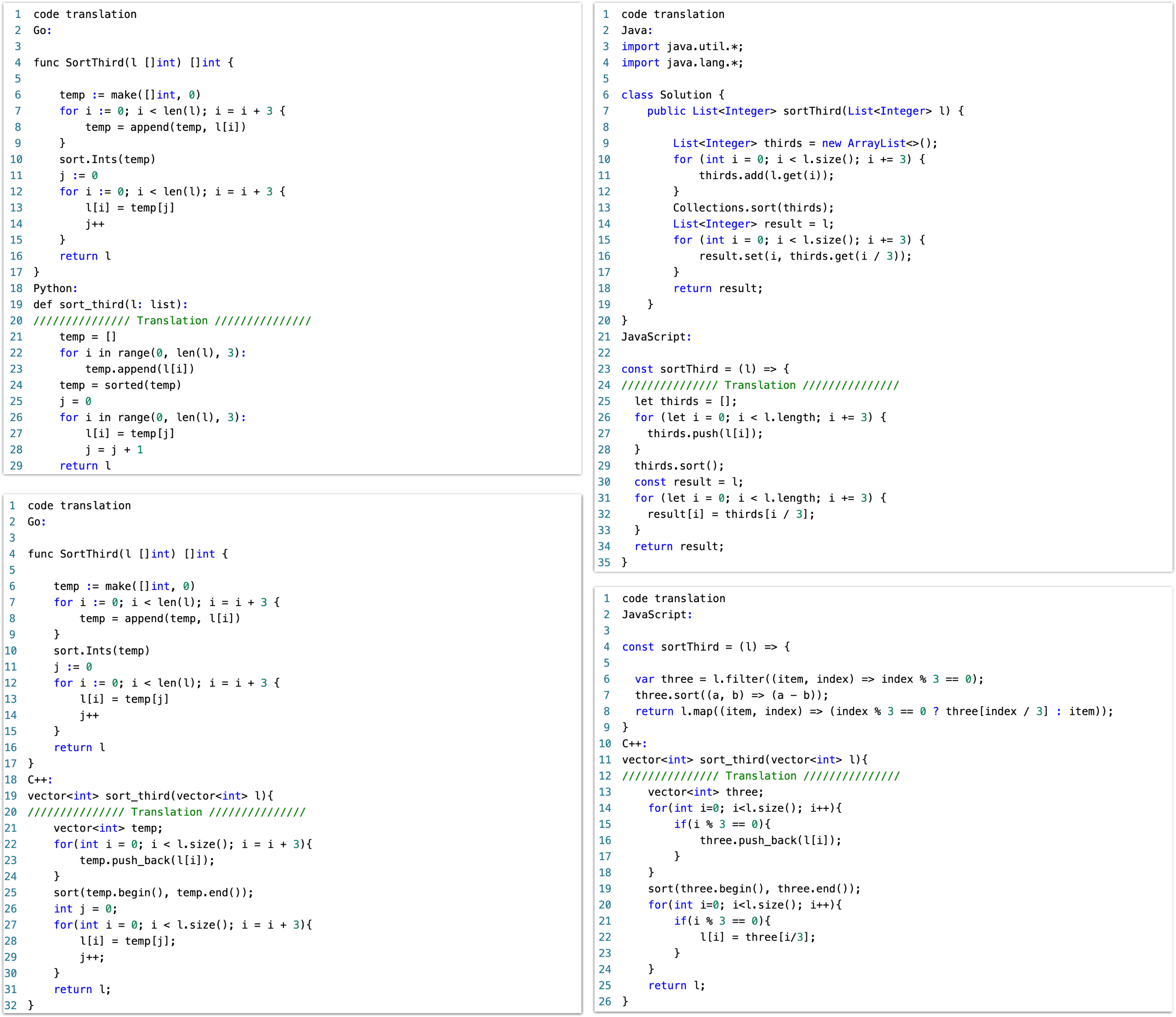}
    \caption{Solutions (Problem 33 in \bench) translated by \name. Prompt and generated codes are separated by the 'Translation' line (added after the generation as an indicator).}
    \label{fig:hx_example_trans_id_33}
    \vspace{-5mm}
\end{figure}

\begin{figure}[htbp]
    \centering
    \includegraphics[width=\textwidth]{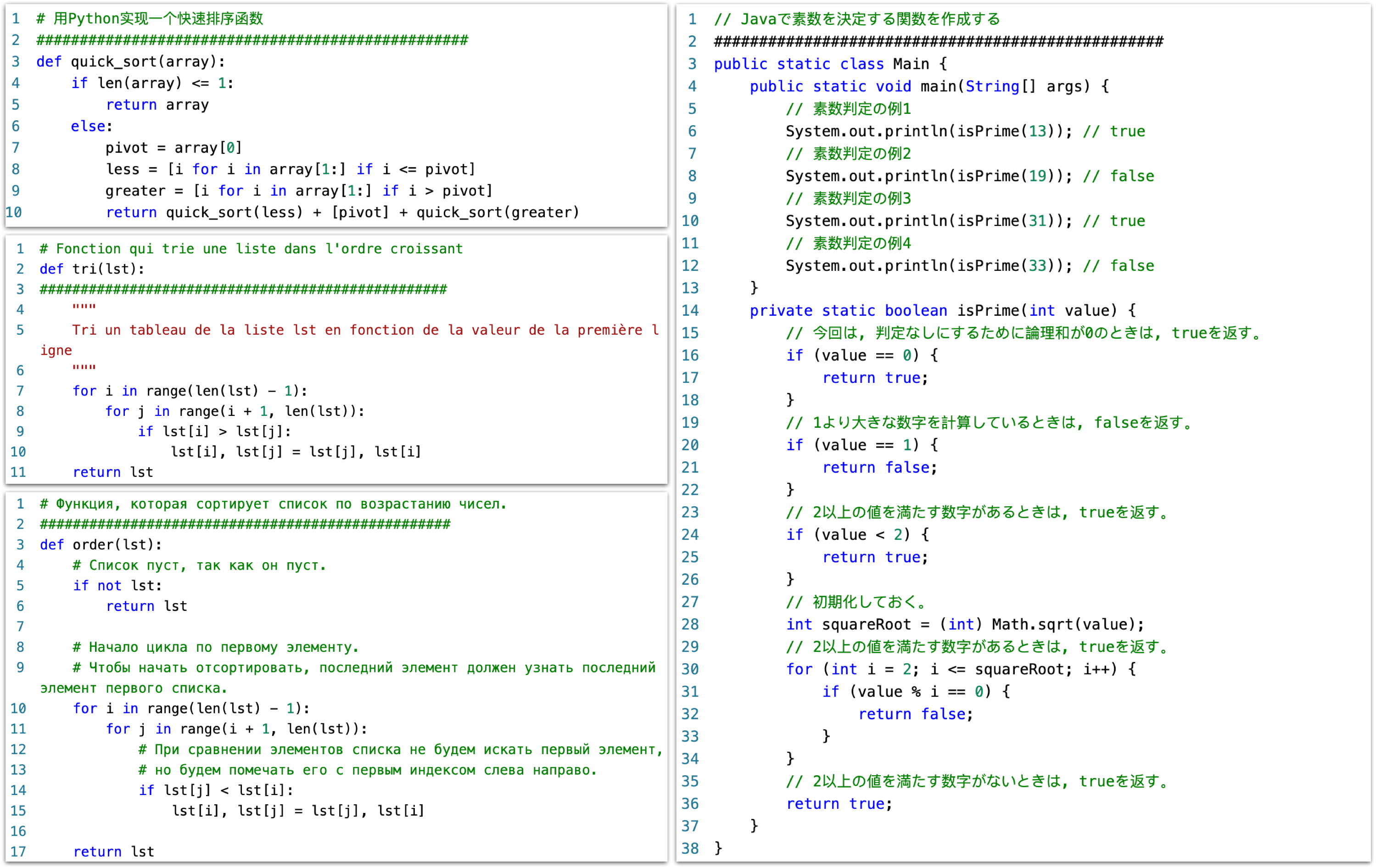}
    \caption{Examples of \name generation with prompts in Chinese, French, Russia and Japanese. Prompt and generated codes are separated by multiple '\#'s (added after the generation as an indicator).}
    \label{fig:example_other_languages}
    \vspace{-5mm}
\end{figure}


\end{document}